\documentclass{article}

\usepackage{microtype}
\usepackage{graphicx}
\usepackage{subcaption}
\usepackage{booktabs} 

\usepackage{hyperref}

\usepackage[preprint]{icml2026}
\usepackage{tabularx} 
\usepackage{graphicx} 
\usepackage{booktabs} 
\usepackage{amsmath}
\usepackage{amssymb}
\usepackage{mathtools}
\usepackage{amsthm}
\usepackage{longtable}
\usepackage{multirow}
\usepackage{enumitem} 
\usepackage{xcolor}      
\usepackage{colortbl}    
\usepackage[table]{xcolor}  
\usepackage{tcolorbox}
\usepackage{verbatim}
\usepackage{tikz}

\usepackage[capitalize,noabbrev]{cleveref}

\theoremstyle{plain}

\theoremstyle{definition}

\theoremstyle{remark}

\icmltitlerunning{HalluSAE:  Detecting Hallucinations in Large Language Models via Sparse Auto-Encoders}

\begin{document}

\twocolumn[
  \icmltitle{HalluSAE:  Detecting Hallucinations in Large Language Models via Sparse Auto-Encoders}

  {\centering \normalsize 
   Boshui Chen, Zhaoxin Fan$^*$, Ke Wang, Zhiying Leng, Faguo Wu, Hongwei Zheng, Yifan Sun, Wenjun Wu \par}
  \vspace{0.15in} 

  \icmlkeywords{Machine Learning, ICML, Hallucination, Interpretability, LLM}
  \vskip 0.3in
]


\begingroup
\renewcommand{\footnotetext}[1]{}
\printAffiliationsAndNotice{}
\endgroup

\begingroup
\renewcommand\thefootnote{}
\footnotetext{
  Boshui Chen, Zhaoxin Fan, Ke Wang, Faguo Wu, and Wenjun Wu are from Beijing Advanced Innovation Center for Future Blockchain and Privacy Computing, School of Artificial Intelligence, Beihang University, Beijing, China. \\
  \hspace*{1.5em}Zhiying Leng is from State Key Laboratory of Virtual Reality Technology and Systems, Beihang University, Beijing, China. \\
  \hspace*{1.5em}Hongwei Zheng is from Beijing Academy of Blockchain and Edge Computing, Beijing, China. \\
  \hspace*{1.5em}Yifan Sun is from Renmin University of China, Beijing, China. \\
  \hspace*{1.5em}$^*$Correspondence to: Zhaoxin Fan $<$zhaoxinf@buaa.edu.cn$>$.
}
\addtocounter{footnote}{-1}
\endgroup

\begin{abstract}
Large Language Models (LLMs) are powerful and widely adopted, but their practical impact is limited by the well-known hallucination phenomenon. While recent hallucination detection methods have made notable progress, we find most of them overlook the dynamic nature and underlying mechanisms of it. To address this gap, we propose \textbf{HalluSAE}, a phase transition-inspired framework that models hallucination as a critical shift in the model’s latent dynamics. By modeling the generation process as a trajectory through a potential energy landscape, HalluSAE identifies critical transition zones and attributes factual errors to specific high-energy sparse features. Our approach consists of three stages: (1) Potential Energy Empowered Phase Zone Localization via sparse autoencoders and a geometric potential energy metric; (2) Hallucination-related Sparse Feature Attribution using contrastive logit attribution; and (3) Probing-based Causal Hallucination Detection through linear probes on disentangled features. Extensive experiments on Gemma-2-9B demonstrate that HalluSAE achieves state-of-the-art hallucination detection performance.
\end{abstract}

\section{Introduction}
\label{sec:intro}

Large Language Models (LLMs) have demonstrated remarkable emergent abilities~\cite{wei2022emergent}, but still suffer from the well-known issue of hallucination, where the model generates content that is plausible yet incorrect~\cite{ji2023survey}. This issue limits the use of LLMs in critical domains such as healthcare~\cite{singhal2023large} and law~\cite{dahl2024large}, raising concerns about their reliability. Therefore, understanding the underlying causes of \mbox{hallucination~\cite{elhage2021mathematical, gao2024scaling}} and conducting effective hallucination detection has become a key research focus.

\begin{figure}[t]
    \centering
    
    \includegraphics[width=\linewidth]{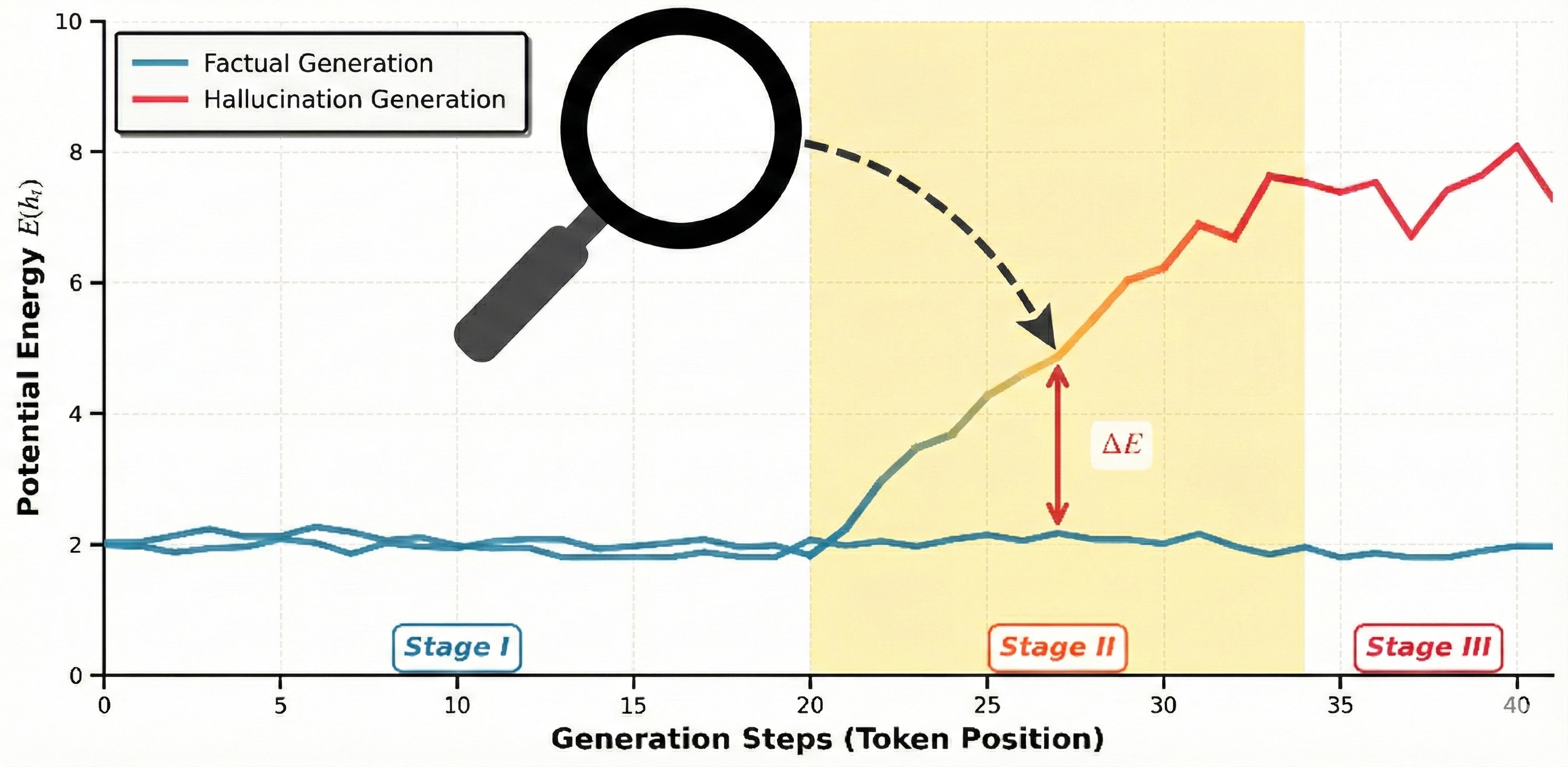}
    \caption{\textbf{Illustration of Phase Transition in LLM's  reasoning trajectories.} 
The trajectories in potential energy space reveal three phases: 
early stability (Phase I), critical transition (Phase II, yellow highlight), and sustained error plateau (Phase III).  Factual generation (blue) maintains low energy throughout, while hallucination (gradient color) undergoes abrupt energy increase ($\Delta E$) during the transition zone, entering a persistent high-energy state.}
    \label{fig:teaser}
    \vspace{-0.2in}
\end{figure}

Mainstream hallucination detection methods are typically divided into black-box and white-box approaches. Black-box methods, like SelfCheckGPT~\cite{manakul2023selfcheckgpt} and CoVe~\cite{dhuliawala2024chain}, rely on output consistency across multiple model runs, but are computationally intensive and lack interpretability. White-box methods instead leverage internal states, using heuristics such as perplexity~\cite{huang2025reppl,friel2023chainpoll}, attention entropy~\cite{li2025language}, or activation statistics. More advanced approaches, like SAPALMA~\cite{azaria2023internal} and ITI~\cite{li2023inference}, use linear probes on the residual stream, while others model errors as constraint satisfaction in attention heads~\cite{yuksekgonul2023attention}. Recent studies~\cite{sriramanan2024llm} have shown that internal-state methods offer significant speed advantages over black-box approaches. Although existing hallucination detection methods have achieved remarkable results, we find that they primarily analyze hallucination from the perspective of static feature representations or final outputs. Consequently, they do not fully consider the dynamic nature of hallucination as it unfolds during the generation process, and often overlook important information about how hallucinations develop over time. Moreover, focusing only on static features makes it easy for individual neurons to mix together many unrelated concepts~\cite{elhage2021mathematical}, which can negatively impact detection performance.

Therefore, this paper proposes a physics-inspired perspective to bridge this gap. Rather than interpreting hallucinations as isolated surface errors, we posit that they reflect a critical phase transition within the model’s latent dynamics (see Fig.~\ref{fig:teaser}). By conceptualizing the generation process as a trajectory over a potential energy landscape, we reveal a distinctive mechanism: factual errors emerge when internal representations, propelled by specific high-energy sparse features, abruptly transition from a low-energy “truth attractor” to a persistent, unstable high-energy plateau. This dynamic viewpoint motivates a key insight: \textit{by monitoring the temporal evolution of latent potential energy in LLM, one can potentially localize phase transition zones that serve as reliable indicators for hallucination detection.}

To this end, we propose a systematic detection framework, \textbf{HalluSAE}, grounded in the phase transition perspective. Our approach follows a coarse-to-fine pipeline comprising three stages: (1) \textit{Potential Energy Empowered Phase Zone Localization}, where we leverage the Gemma Scope Sparse Autoencoder (SAE) to disentangle the residual stream and introduce a geometric potential energy metric, allowing us to efficiently localize “phase transition zones” by identifying layers with exponential energy growth and narrowing the search space to a few critical transition points; (2) \textit{Hallucination-related Sparse Feature Attribution}, in which we apply Contrastive Direct Logit Attribution (DLA) within these high-energy regions to quantify the contributions of individual sparse features to incorrect outputs, thereby precisely isolating “hallucination-inducing features” responsible for specific error patterns; and (3) \textit{Probing-based Causal Hallucination Detection}, where we train linear probes on the selected sparse features to enable accurate and efficient inference-time hallucination detection, establishing a causal link between internal model dynamics and output errors. Collectively, HalluSAE offers a targeted, interpretable, and scalable solution for hallucination detection by explicitly modeling the dynamic processes underlying factual errors in large language models. 

To assess the effectiveness and generalizability of our approach, we conduct experiments on Gemma-2-9B using both in-distribution (HaluEval) and out-of-distribution (TriviaQA) benchmarks, where our method consistently achieves superior hallucination detection performance over existing baselines. The main contributions of this paper are summarized as follows:

\begin{itemize}
\item We introduce HalluSAE, a framework that approaches hallucination from a phase transition perspective, enabling dynamic and process-aware hallucination detection rather than relying solely on static representations.
\item We propose Potential Energy Empowered Phase Zone Localization, Hallucination-related Sparse Feature Attribution, and Probing-based Causal Hallucination Detection in HalluSAE to precisely identify and attribute hallucination-inducing features.
\item We conduct comprehensive experiments, which demonstrate that our method achieves new state-of-the-art results on in-distribution (HaluEval) tasks, and  out-of-distribution (TriviaQA) benchmarks.
\end{itemize}

\section{Related Work}
\label{sec:related}

\noindent \textbf{Hallucination in Large Language Models.} Despite the impressive emergent abilities of LLMs~\cite{wei2022emergent}, hallucinations remain a persistent challenge, especially in high-stakes domains such as healthcare~\cite{singhal2023large} and law~\cite{dahl2024large}. Recent taxonomies categorize these errors as input-conflicting or fact-conflicting, attributing them to data divergence or the compounding of early generation mistakes~\cite{zhang2023siren, huang2025survey}. However, while these classifications are well-studied, the mechanistic understanding of \textit{how} and \textit{why} hallucinations dynamically arise during inference is still lacking \cite{orgad2024llms, zhang2025law}. Most existing methods treat hallucination as a static phenomenon based on final output or representations \cite{ji2023survey, huang2025survey}; in contrast, this work analyzes hallucination from a dynamical perspective for its detection.

\noindent \textbf{Hallucination Detection Methods.} Existing hallucination detection strategies fall into black-box and white-box categories. Black-box methods rely on output consistency checks~\cite{manakul2023selfcheckgpt, dhuliawala2024chain}, {external LLM judges~\cite{benkirane2024machine}, or Bayesian sequential estimation~\cite{wang2023hallucination}}, but are often computationally costly and lack interpretability. {Zero-shot attention methods like AGSER~\cite{liu2025attention} also require multiple inference passes, increasing latency.} White-box methods leverage internal states. {Early work by Xu et al.~\cite{xu2023understanding} pioneered model introspection in NMT. Recently, HARP~\cite{hu2025harp} and GSP~\cite{noel2025graph} utilized SVD projections and graph spectral energy, respectively, to detect errors.} However, these {dense or topological approaches} often suffer from neuron polysemanticity~\cite{elhage2021mathematical}, making it difficult to isolate causal features. In contrast, we take a sparse and dynamic perspective, utilizing SAEs to precisely localize hallucination causes.

\noindent \textbf{Sparse Autoencoders and Their Application.} 
Sparse Autoencoders (SAEs) have emerged as a powerful tool for interpretability, decomposing dense activations into human-interpretable features~\cite{bricken2023monosemanticity, cunningham2023sparse, lieberum2024gemma}. Recent studies have demonstrated their versatility in uncovering functional circuits~\cite{marks2024sparse} and enhancing privacy~\cite{frikha2025privacyscalpel}. In hallucination research, SAFE~\cite{abdaljalil2025safe}, SSL~\cite{hua2025steering}, SAVE~\cite{park2025save}, and RAGLens~\cite{xiong2025toward} utilize SAEs for query enrichment, steering, or detecting faithfulness failures in RAG. However, these methods either focus on RAG-specific scenarios, require task-specific feature selection, or treat generation as static activation without modeling the \textit{dynamic phase transition} process. In contrast, we introduce a geometric potential energy framework that tracks temporal evolution and systematically localizes causal features, making this the first work leveraging dynamic SAE geometry for general factual hallucination detection.



\section{Pre-analysis on Hallucination From the Phase Transition View}
\label{sec:exploration}

As discussed above, this work analyzes hallucination in LLMs from a phase transition perspective and proposes a  hallucination detection framework based on this view. Before introducing our HalluSAE method, we first present the key analytical tools and some preliminary  findings. Specifically, to analyze hallucination in LLMs, we primarily leverage the two following tools: .

\noindent\textbf{Sparse Autoencoders (SAEs).}
SAEs address the problem of mixed semantic representations in dense activations by mapping the $d_{\text{model}}$-dimensional residual stream to a much higher-dimensional, sparse space ($d_{\text{SAE}} \gg d_{\text{model}}$), where each dimension is encouraged to capture a single semantic concept~\cite{cunningham2023sparse, bricken2023monosemanticity}.
In this work, we adopt the open-source Gemma Scope SAE~\cite{lieberum2024gemma}, which decomposes the residual stream as a weighted sum of sparse features:
\begin{equation}
    r_l^t \approx \sum_{i=1}^{d_{\text{SAE}}} s_i \cdot W_{\text{dec}}[i, :]
\end{equation}
where $s_i = \text{JumpReLU}(W_{\text{enc}} \cdot r_l^t + b_{\text{enc}})[i]$ denotes the activation of the $i$-th sparse feature.
We set $d_{\text{SAE}} = 131{,}072$ (36.6$\times$ overcomplete) and use $L_0 \approx 30$ for a balance between sparsity and interpretability.
The JumpReLU activation preserves activation strength while maintaining sparsity, thus better capturing the underlying semantics compared to traditional L1-SAEs.

\noindent\textbf{Geometric Potential Energy (GPE).}
From a dynamical systems perspective, we quantify the deviation from a stable, factual state using Geometric Potential Energy (GPE), defined as the squared Euclidean distance from the ``truth attractor'' in SAE feature space~\cite{marks2023geometry, zou2023representation}:
\begin{equation}
    E(l, t) = \|\text{SAE}(r_l^t) - \mu_{\text{truth}}\|_2^2
    \label{eq:gpe}
\end{equation}
where $\mu_{\text{truth}}$ is the centroid of SAE features for all factual samples.
This metric captures both the direction and magnitude of deviation, amplifies significant departures (consistent with phase transition phenomena), and focuses on semantic-level differences thanks to the disentangling property of SAEs.

\begin{figure}[t]
    \centering
    
    \includegraphics[width=0.95\columnwidth]{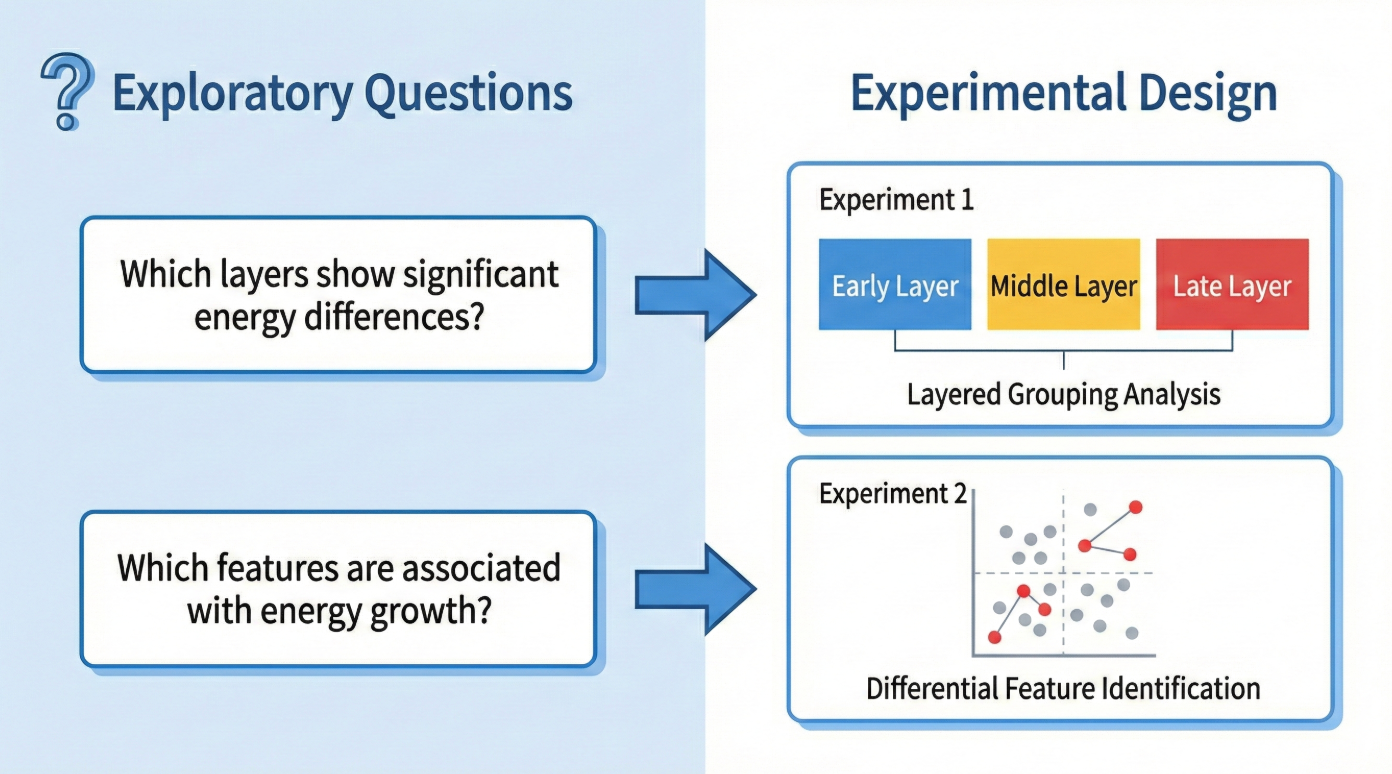}
    \caption{\textbf{Illustration of the Exploratory Experimental Design.} We conduct two complementary experiments: \textbf{Exp 1} investigates layer-wise energy distribution by dividing 42 layers into Early/Middle/Late groups and comparing GPE differences; \textbf{Exp 2} identifies microscopic feature-level contributions by analyzing differential features that exhibit high activation in hallucination samples but low activation in factual samples.}
    \label{fig:exp_design}

\end{figure}

\begin{figure}[t]
    \centering
    
    \includegraphics[width=\columnwidth]{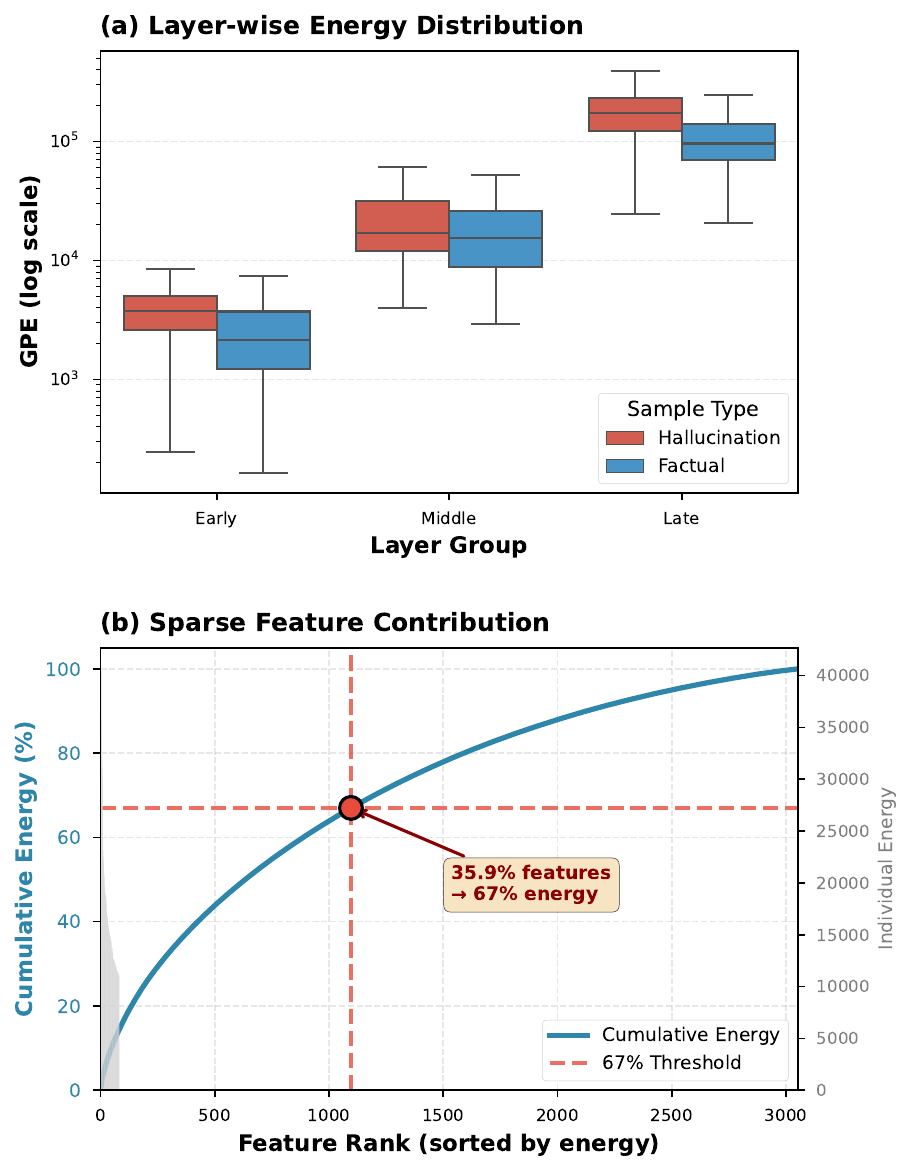}
    \caption{\textbf{Experimental Results on Hallucination Dynamics.}
    (a) \textbf{Layer-wise energy distribution (grouped analysis).}
    Grouped box plots reveal that hallucination samples' GPE exhibits significant escalation from Early to Late layer groups (***$p<0.001$). Error bars indicate 95\% confidence intervals (bootstrap, $n{=}200$ per group).
    (b) \textbf{Sparse feature contribution analysis.}
    Cumulative energy contribution curve demonstrates that a small fraction of differential features account for the majority of total energy increase, exhibiting a highly skewed distribution. Gray bars represent individual feature energy contributions.}
    \label{fig:exploratory}
    \vspace{-0.25in}
\end{figure}

Using both analytical tools, as shown in Fig.~\ref{fig:exp_design}, we first conduct a series of controlled experiments to pre-analyze the internal dynamics of LLM reasoning. Our experimental results illustrated in Fig.~\ref{fig:exploratory} reveal several key patterns. First, layer-wise energy analysis (Fig.~\ref{fig:exploratory}a) shows that the energy difference between hallucination and factual samples is negligible in the early layers, increases steadily in the middle layers, and surges sharply in the late layers—exhibiting a clear phase transition~\cite{meng2022locating, geva2021transformer}. Statistical analysis (Cohen’s $d = 1.64$, $p < 0.001$) further confirms that hallucinations arise within specific layer intervals, rather than being uniformly distributed across the network. Second, at the feature level (Fig.~\ref{fig:exploratory}b), we observe that only a small subset of sparse features—those highly activated in hallucination cases—are responsible for the majority of the energy increase. This heavy-tailed distribution indicates that hallucinations are driven by a few high-energy features, rather than by diffuse, low-level noise. In summary, our analysis yields two main findings:
\begin{itemize}
    \item \textbf{Finding 1}: Hallucination emerges as a dynamic process during generation, with clear phase transitions occurring at specific layer intervals rather than manifesting as a static property throughout the network.
    \item \textbf{Finding 2}: Hallucination is predominantly driven by a small subset of high-impact sparse activations, rather than by broad or diffuse noise, highlighting the critical role of a few key features in triggering the transition to erroneous outputs.
\end{itemize}

These two findings motivate our approach: \textit{by dynamically monitoring potential energy changes in the sparse feature space, we can effectively detect phase transitions that indicate hallucination.} Next, we introduce our motivated approach HalluSAE in detail.



\section{Method}
\label{sec:method}

Given an input prompt, our goal is to predict whether the model's output will contain hallucinations before generation is complete—by analyzing activations from selected layers inside the LLM. Motivated by our earlier findings, we propose HalluSAE, a three-stage framework grounded in the dynamics of sparse feature activations and phase transitions. 

HalluSAE operates as follows:
(1) Potential Energy Empowered Phase Zone Localization leverages sparse autoencoders and a geometric potential energy metric to identify critical layer intervals where phase transitions indicative of hallucination occur.
(2) Hallucination-related Sparse Feature Attribution uses contrastive logit attribution to pinpoint a small set of sparse features most responsible for hallucination within these layers.
(3) Probing-based Causal Hallucination Detection conducts linear probing on disentangled features to detect and validate causal relationships between feature activation patterns and hallucination outcomes. In the following subsections, we detail the methodology and implementation of each stage.

\subsection{Stage I: Potential Energy Empowered Phase Zone Localization}
\label{subsec:stage1}

The primary goal of this module is to identify which layers within the model are most susceptible to hallucination, by detecting phase changes through differences in Geometric Potential Energy (GPE) between hallucination and factual samples. By quantifying how the GPE gap evolves across layers, we aim to pinpoint the specific regions in the network where the internal state shifts most sharply toward hallucination. The operations are as follows:

For each layer $\ell$, we compute the average GPE difference as
\begin{equation}
\Delta E_\ell = \mathbb{E}_{x \in \mathcal{H}}[E_\ell(x)] - \mathbb{E}_{x \in \mathcal{F}}[E_\ell(x)],
\end{equation}
where $\mathcal{H}$ and $\mathcal{F}$ denote the sets of hallucination and factual samples, respectively. To characterize the dynamics of this difference, we define the relative growth rate
\begin{equation}
\gamma_\ell = \frac{\Delta E_\ell - \Delta E_{\ell-1}}{\Delta E_{\ell-1}} \times 100\%.
\end{equation}
A stable or oscillatory $\gamma_\ell$ suggests the model remains in a consistent state, while a sequence of sustained positive growth indicates a transition toward hallucination.

To robustly identify the onset of this transition, we locate the layer interval [$\ell_{\mathrm{start}}$, $\ell_{\mathrm{end}}$] starting from the earliest layer $\ell_{\mathrm{start}}$ that exhibits three consecutive positive growth rates,
\begin{equation}
\ell_{\mathrm{start}} = \min \left\{ \ell \mid \gamma_\ell > 0, \gamma_{\ell+1} > 0, \gamma_{\ell+2} > 0 \right\}.
\end{equation}
We then define the endpoint $\ell_{\mathrm{end}}$ as the layer where $\Delta E_\ell$ reaches its post-onset maximum,
\begin{equation}
\ell_{\mathrm{end}} = \arg\max_{\ell \geq \ell_{\mathrm{start}}} \Delta E_\ell.
\end{equation}

This process allows us to automatically detect the ``phase transition zone''—the interval in which the model's internal state shifts most dramatically toward hallucination—by analyzing potential energy dynamics. The identified zone provides a principled basis for focusing subsequent fine-grained analysis on the most vulnerable layers for hallucination emergence.

\subsection{Stage II: Hallucination-related Sparse Feature Attribution}
\label{subsec:stage2}

After localizing the layers most vulnerable to hallucination, the core challenge is to understand which internal mechanisms within these layers are responsible for the phenomenon. Due to the polysemanticity and entanglement present in raw model activations, direct analysis is often unreliable. We therefore employ the sparse and overcomplete representations provided by the SAE as a principled basis for disentangling and attributing semantic effects in the network. The interpretability and sparsity of SAE features allow for precise tracing of how specific activations influence model outputs.

Specifically, within the identified phase transition layers, we aim to isolate the subset of SAE features that causally drive hallucination~\cite{elhage2021mathematical, wang2022self}. For each feature $i$, we define its direct logit attribution to a target token $t$ as
\begin{equation}
\mathcal{A}_i^{(t)} = \left( s_i \, \mathbf{w}_{\mathrm{dec}}^{(i)} \right)^\top \mathbf{W}_{\mathrm{U}}[:, t],
\end{equation}
where $s_i$ is the feature activation, $\mathbf{w}_{\mathrm{dec}}^{(i)}$ is the SAE decoder direction, and $\mathbf{W}_{\mathrm{U}}$ is the unembedding matrix. This quantifies the linear pathway from a sparse feature to its impact on the model's output space.

Yet, hallucination generation is inherently a contrastive process: it depends not only on increasing the likelihood of the incorrect token but also on suppressing the correct one~\cite{li2023inference}. To capture this, we introduce the Contrastive Direct Logit Attribution (C-DLA), defined as
\begin{equation}
\mathcal{C}_i = s_i \cdot \left( \mathbf{w}_{\mathrm{dec}}^{(i)} \right)^\top \left( \mathbf{W}_{\mathrm{U}}[:, t_{\mathrm{wrong}}] - \mathbf{W}_{\mathrm{U}}[:, t_{\mathrm{correct}}] \right).
\end{equation}
Here, the vector $\Delta \mathbf{v} = \mathbf{W}_{\mathrm{U}}[:, t_{\mathrm{wrong}}] - \mathbf{W}_{\mathrm{U}}[:, t_{\mathrm{correct}}]$ represents the semantic direction from the correct answer to the hallucinated answer in output space. A positive value of $\mathcal{C}_i$ indicates that feature $i$ both elevates the hallucinated token and suppresses the factual one, thus promoting hallucination.

For each prompt, we compute $\mathcal{C}_i$ for all features in the critical layers, rank them by magnitude, and we select the top 100 features (0.076\% of the 131,072-dimensional space) as the critical set. This procedure yields a compact collection of features that are not merely correlated with hallucination, but mechanistically responsible for shifting model outputs away from factuality. These identified features serve as the analytical and causal substrate for detection and intervention in the next stage.

\subsection{Stage III: Probing-based Causal Hallucination Detection}
\label{subsec:stage3}

Building on the identification of critical SAE features within the phase transition zone, the final step is to leverage these features for hallucination detection. The objective is to determine, for a given prompt and its candidate output, whether the activation pattern across this feature set reliably signals the presence of hallucination.

In particular, for each sample $x$, let $\mathbf{s}_{\mathcal{T}}(x) \in \mathbb{R}^{|\mathcal{T}|}$ denote the concatenated activations of the selected feature set $\mathcal{T}$ across all layers in the identified zone. Rather than relying on granular error categorization, we adopt a unified detection strategy. We construct the critical feature set $\mathcal{T}$ by selecting the top-ranked sparse features based on their average C-DLA magnitude across the entire training dataset. A single logistic regression Probe is then trained on these high-impact features~\cite{alain2016understanding} to distinguish between hallucination and factual generations. This holistic approach leverages the semantic disentanglement of SAEs, enabling a compact feature set to capture diverse hallucination patterns---ranging from numerical errors to entity substitutions---without requiring explicit type-specific engineering.

The detection model is formulated as an $\ell_1$-regularized logistic regression, optimizing
\begin{equation}
\min_{\boldsymbol{\theta}} \ \frac{1}{N} \sum_{i=1}^N \log\left(1 + \exp(-y_i \, \boldsymbol{\theta}^\top \mathbf{s}_{\mathcal{T}}(x_i))\right) + \lambda \|\boldsymbol{\theta}\|_1,
\end{equation}
where $y_i \in \{+1, -1\}$ is the hallucination label, and $\lambda$ is selected by cross-validation. This approach yields a sparse, interpretable classifier in which nonzero coefficients directly indicate the contributions of individual features to hallucination prediction.

During inference, for any input, the model extracts the activation pattern across $\mathcal{T}$, and outputs a hallucination probability. Samples exhibiting activation signatures similar to known hallucinations are thus flagged in advance of output generation, enabling proactive identification.


\section{Experiments}
\label{sec:experiments}

\subsection{Experimental Setup}
\label{subsec:exp_setup}

\begin{table*}[t]
\centering
\caption{\textbf{Comparison Results of Hallucination Detection Performance Across Benchmarks.} We compare our method against eight representative baselines spanning four paradigms. Results are grouped by in-distribution (HaluEval) and out-of-distribution (TriviaQA) settings. Best results are highlighted in gray. All metrics are reported as percentages, with AUC as the primary ranking metric.}
\label{tab:detection_performance_wide}
\small
\setlength{\tabcolsep}{8pt}
\begin{tabularx}{\textwidth}{@{}llXXXXXX@{}}
\toprule
& & \multicolumn{3}{c}{\textbf{In-Distribution (HaluEval)}} & \multicolumn{3}{c}{\textbf{Out-of-Distribution (TriviaQA)}} \\
\cmidrule(lr){3-5} \cmidrule(lr){6-8}
\textbf{Category} & \textbf{Method} & \textbf{AUC} $\uparrow$ & \textbf{Acc.} $\uparrow$ & \textbf{Recall} $\uparrow$ & \textbf{AUC} $\uparrow$ & \textbf{Acc.} $\uparrow$ & \textbf{Recall} $\uparrow$ \\
\midrule
\multirow{2}{*}{Uncertainty} 
  & \textbf{LN-Entropy} & 54.93 & 56.50 & 65.12 & 70.94 & 65.75 & 66.50 \\
  & \textbf{Semantic Entropy} & 62.77 & 67.98 & 55.45 & 71.85 & 69.21 & 71.88 \\
\midrule
\multirow{2}{*}{Consistency} 
  & \textbf{Lexical Similarity} & 64.25 & 62.50 & 59.00 & 65.40 & 64.25 & 53.50 \\
  & \textbf{SelfCheckGPT} & 63.78 & 64.12 & 56.23 & 63.53 & 61.51 & 57.12 \\
\midrule
\multirow{2}{*}{Internal State} 
  & \textbf{EigenScore} & 60.53 & 59.50 & 77.13 & 61.51 & 60.25 & 66.50 \\
  & \textbf{MHAD} & 64.18 & 63.32 & 66.20 & 61.01 & 60.75 & 76.51 \\
\midrule
\multirow{2}{*}{Supervised} 
  & \textbf{SAPLMA} & 79.14 & 75.34 & 75.19 & 64.33 & 59.50 & 55.12 \\
  & \textbf{HaloScope} & 84.37 & 78.62 & 75.96 & 78.94 & 72.21 & 72.83 \\
\midrule
\rowcolor{gray!15}
\textbf{Ours} & \textbf{HalluSAE} & \textbf{92.86} & \textbf{82.81} & \textbf{80.88} & \textbf{80.44} & \textbf{76.20} & \textbf{75.60} \\
\bottomrule
\end{tabularx}
\vspace{-0.1in}
\end{table*}

\textbf{Datasets.}
We evaluate our method on two complementary benchmarks. For in-distribution (ID) evaluation, we use the HaluEval dataset~\cite{li2023halueval}, consisting of 1,800 samples (900 hallucination, 900 factual) with 80\% allocated for training (1,440 samples) and 20\% for testing (360 samples). All samples undergo dual annotation: automatic verification by GPT-4o followed by human expert review. The dataset is carefully balanced across generation length, error types (50\% numerical, 50\% entity), and question difficulty. For out-of-distribution (OOD) evaluation, we construct a TriviaQA test set containing 4,000 samples (2,000 hallucination, 2,000 factual) to assess zero-shot cross-domain generalization. Compared to HaluEval, TriviaQA features more colloquial phrasing, broader knowledge coverage spanning geography, history, people, and organizations, and greater diversity in question formats.

\textbf{Baselines.}
We compare against eight representative detection methods spanning four paradigms. Uncertainty-based approaches include LN-Entropy~\cite{manakul2023selfcheckgpt, xiao2021hallucination} and Semantic Entropy~\cite{farquhar2024detecting}. Consistency-based methods comprise Lexical Similarity~\cite{manakul2023selfcheckgpt, wang2022self} and SelfCheckGPT~\cite{manakul2023selfcheckgpt}. Internal state-based approaches feature EigenScore~\cite{chen2024inside} and MHAD~\cite{zhang2025detecting}. Supervised probing methods include SAPLMA~\cite{azaria2023internal} and HaloScope~\cite{du2024haloscope}. Detailed implementation configurations for all baselines are provided in Appendix~\ref{app:baseline_config}.

\textbf{Experimental Details.} 
All experiments are conducted on Gemma-2-9B (42 layers) equipped with Gemma Scope Sparse Autoencoders~\cite{lieberum2024gemma}. The SAE configuration includes 131,072 feature dimensions (36.6$\times$ overcomplete), JumpReLU activation function, and average sparsity $L_0 \approx 30$ across layers. Stage I localizes the phase transition zone by identifying layers with exponential growth in geometric potential energy, yielding layers 23--35 (L23--35) as the critical interval. In Stage II, Contrastive Direct Logit Attribution (C-DLA) is applied within this window to select the top-100 features exhibiting the highest causal influence on hallucination. We report Area Under the ROC Curve (AUC) as the primary metric for assessing ranking capability, supplemented by accuracy, recall, and specificity for comprehensive performance characterization. 

\subsection{Main Results}
\label{subsec:main_results}

\textbf{Quantitative Comparison.} Table~\ref{tab:detection_performance_wide} summarizes performance across all methods. HalluSAE attains an AUC of 0.9286 on the in-distribution test set and 0.8044 on out-of-distribution TriviaQA, establishing new state-of-the-art results in both regimes. Unsupervised baselines display limited efficacy: the strongest (Semantic Entropy) achieves 0.7185 AUC on OOD data, while HalluSAE surpasses this by 11.9\% (0.8044 vs.\ 0.7185), highlighting the insufficiency of surface-level statistical cues for capturing the causal dynamics of hallucination. Consistency-based approaches perform worse still, with SelfCheckGPT reaching only 0.6353 OOD AUC—a 23.0\% deficit relative to HalluSAE. Among supervised baselines, HaloScope represents the previous state-of-the-art, utilizing dense hidden states (3,584 dimensions) across multiple layers. HalluSAE outperforms HaloScope by 10.1\% on in-distribution data (0.9286 vs.\ 0.8437) and by 1.9\% on OOD data (0.8044 vs.\ 0.7894). This improvement is attributable to three core advantages of sparse autoencoder features: (i) semantic disentanglement in the 131k-dimensional sparse space mitigates the polysemanticity of dense activations, where individual neurons encode multiple unrelated concepts; (ii) C-DLA enables precise, mechanistic attribution of hallucination to individual features, rather than diffuse patterns over thousands of variables; and (iii) semantic-level representations demonstrate superior robustness to domain shift compared to raw activation statistics. Notably, these gains are achieved using only 100 sparse features—a 35.8$\times$ reduction compared to HaloScope—underscoring the efficiency and interpretability of the approach.

\begin{figure}[ht]
\centering

\includegraphics[width=\columnwidth]{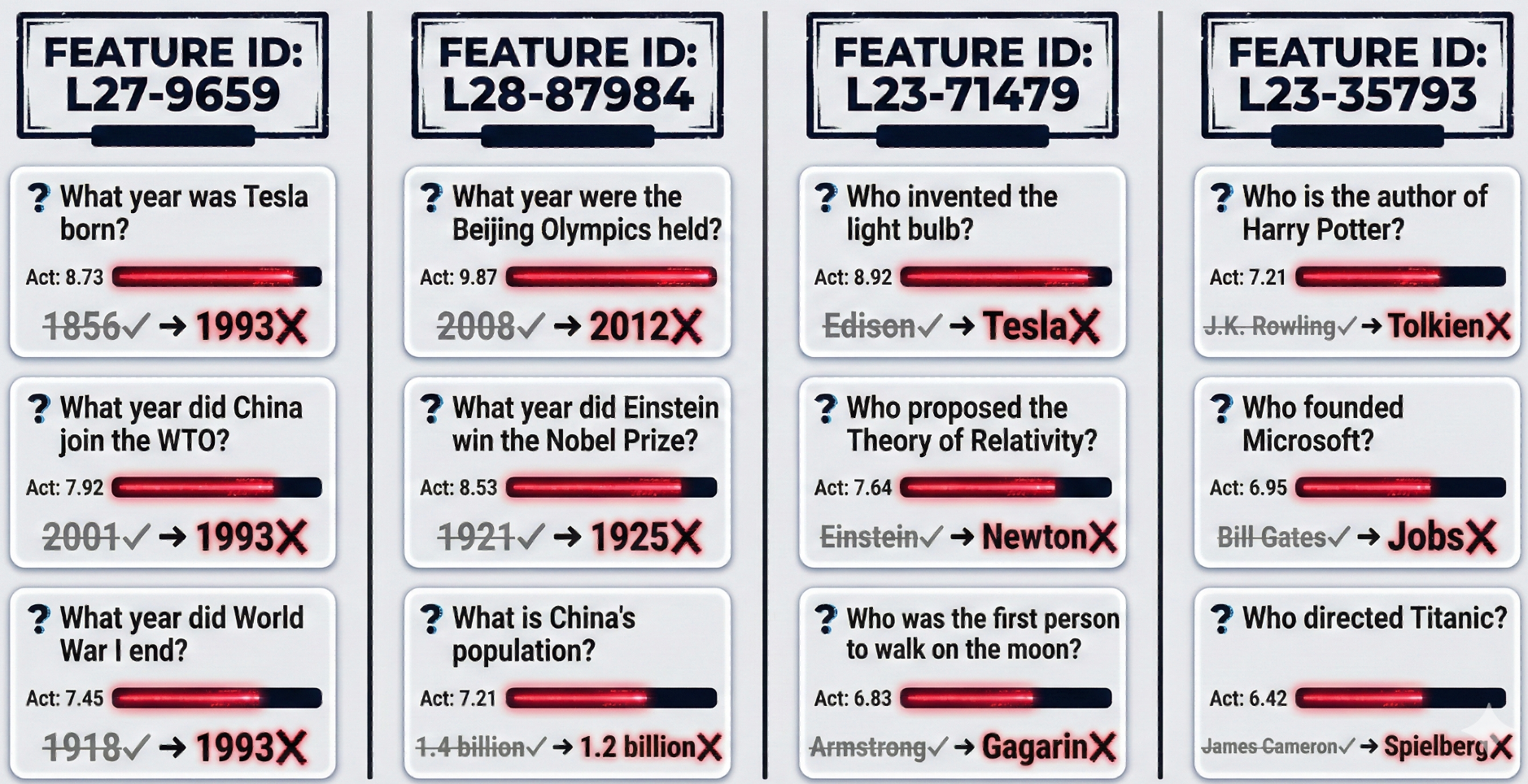}
\caption{\textbf{Examples of Samples with High-Activation on Identified Sparse Features.} Each feature exhibits stable and interpretable error patterns: L27-9659 shows fixed year substitution (always outputs ``1993''), L28-87984 demonstrates numerical shift patterns (+4 years, -0.2 billion), L23-71479 drives domain-specific entity confusion (Tesla/Edison, Newton/Einstein), and L24-35793 triggers cross-domain celebrity substitution (Tolkien/Rowling, Jobs/Gates).}
\label{fig:case_study}
\vspace{-0.2in}
\end{figure}

\textbf{Qualitative Comparison.}  To assess the interpretability and stability of the features uncovered by our approach, we conduct an in-depth analysis of the top-50 features ranked by C-DLA score. We select features according to two stringent criteria: (1) high-activation samples for a given feature are exclusively hallucination cases, and (2) the feature exhibits a significant activation gap between hallucinated and factual samples. Applying these criteria, we identify four representative high-purity features (see Table~\ref{tab:selected_features}). Fig.~\ref{fig:case_study} visualizes the highest-activation samples associated with each feature. Our analysis reveals that samples activating a particular feature consistently display distinct error modes: for example, features L27-9659 and L28-87984 are linked to numerical substitution errors, while features L23-71479 and L24-35793 correspond to entity replacement phenomena. This strong correspondence between individual features and specific hallucination types demonstrates that these features capture systematic, semantically meaningful deviations—rather than reflecting mere statistical noise. These findings highlight three key properties of our approach. First, there is a robust feature-to-pattern correspondence: each high C-DLA feature is associated with a stable, interpretable hallucination error mode, supporting the view that hallucinations are driven by specific neural mechanisms rather than random fluctuations. Second, the use of sparse autoencoder features yields genuine semantic interpretability, in contrast to dense hidden state representations; extracted features map transparently onto functions such as fixed value anchoring, numerical shift, and entity confusion. Third, the semantic clarity of these features opens the door to targeted causal interventions, suggesting promising directions for hallucination mitigation via SAE-based manipulation.

\begin{table}[h]
\centering
\caption{\textbf{Identified High-Purity Hallucination-Driving Features}}
\label{tab:selected_features}
\small
\scalebox{0.90}{
\begin{tabular}{@{}lccc@{}}
\toprule
\textbf{Feature ID} & \textbf{Hal./Fact. Ratio} & \textbf{Top-10 Purity} & \textbf{Mean Activation} \\
\midrule
L27-9659 & 22.92 & 10/10 (100\%) & 8.03 \\
L28-87984 & 81.52 & 10/10 (100\%) & 8.54 \\
L23-71479 & 16.57 & 9/10 (90\%) & 7.80 \\
L24-35793 & 12.41 & 10/10 (100\%) & 6.86 \\
\bottomrule
\end{tabular}
}
\vspace{-0.1in}
\end{table}

\subsection{Ablation Study}
\label{subsec:ablation}

\begin{figure}
\centering

\includegraphics[width=\columnwidth]{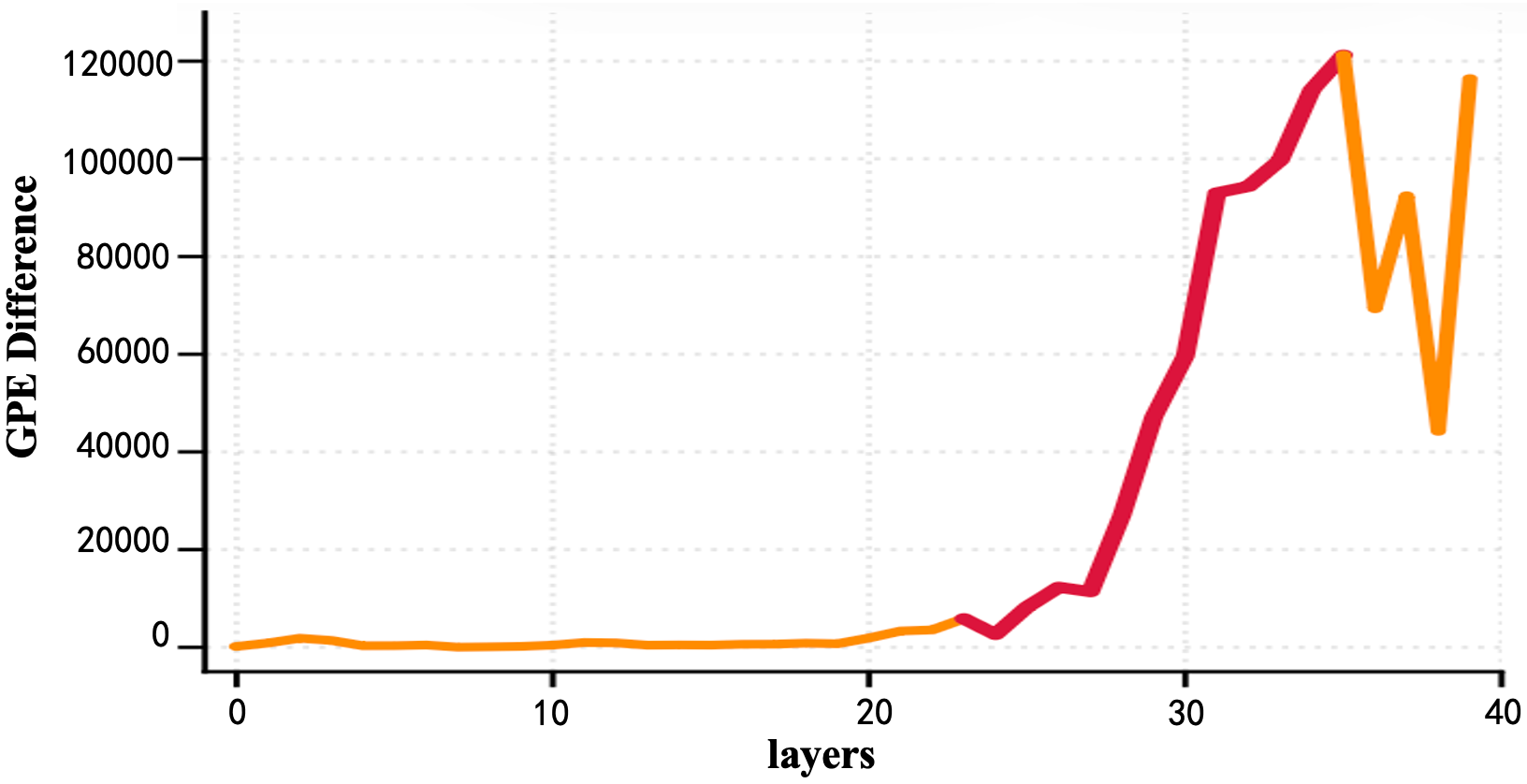}
\vspace{-0.1in}
\caption{\textbf{Illustration of Layer-wise Energy Difference Evolution Across All 42 Layers.} Geometric Potential Energy (GPE) difference (Hallucination - Factual) reveals three distinct phases: stable period (L0--22, near-zero difference with random fluctuations), transition zone (L23--35, sharp 20.7-fold escalation highlighted by shaded region), and plateau period (L36--41, sustained high-energy state). Error bars represent 95\% confidence intervals (bootstrap, $n$=900 per group).}
\label{fig:energy_evolution_full}
\vspace{-0.25in}
\end{figure}

\textbf{Ablation Study on the Potential Energy Empowered Phase Zone
Localization Stage.}
To assess the necessity of precise phase zone localization, we compare five layer selection strategies while fixing the feature dimension at 100. As shown in Table~\ref{tab:layer_ablation_compact}, our energy-identified transition zone (L23--35) achieves optimal results, with an in-distribution AUC of 0.93 and an out-of-distribution AUC of 0.80. Remarkably, this matches the performance of using all 42 layers (420 features), but with a $4.2\times$ reduction in feature count. In contrast, early layers (L0--12) fail to capture hallucination-relevant signals (AUC: 0.81 ID, 0.73 OOD), while late layers (L29--41) yield only intermediate performance (AUC: 0.88 ID, 0.78 OOD), likely reflecting downstream consequences rather than the underlying phase transition process. Randomly selected layers further underperform (AUC: 0.83 ID, 0.76 OOD), underscoring that energy-based localization captures genuine causal structure rather than arbitrary layer boundaries. To further illustrate the phase transition mechanism, Fig.~\ref{fig:energy_evolution_full} depicts the layer-wise evolution of GPE differences between hallucination and factual samples across all 42 layers. The trajectory reveals three distinct phases: a stable period (L0--22) characterized by near-zero energy difference and random fluctuations, indicating similar representational dynamics for hallucinated and factual samples in early processing; a transition zone (L23--35) marked by a dramatic 20.7-fold escalation in energy (from 5,862 at L23 to 121,245 at L35), representing a qualitative shift, as confirmed by a large effect size (Cohen's $d=1.64$, $p<0.001$) between stable and transition phases; and a plateau phase (L36--41) with persistently high but stable energy levels (Cohen's $d=0.12$ vs.\ transition zone, $p=0.342$), indicating entry into a sustained deviated state.

\begin{table*}[t]
\centering
\caption{\textbf{Ablation Results on Layer Selection Strategy.} Comparison of detection performance across different layer selection strategies using a fixed feature budget of 100 dimensions. The transition zone (L23--35) achieves optimal performance with minimal features, demonstrating the effectiveness of our energy-based localization method.}
\label{tab:layer_ablation_compact}
\small
\setlength{\tabcolsep}{6pt}
\begin{tabular}{@{}lcccccc@{}}
\toprule
\textbf{Strategy} & \textbf{Layer Range} & \textbf{\# Layers} & \textbf{\# Feat.} & \textbf{ID AUC} & \textbf{OOD AUC} & \textbf{$\Delta$ vs Ours} \\
\midrule
Random-13 & Random 13 layers & 13 & 100 & 0.83 & 0.76 & -4.0\% \\
Early-13 & L0--12 (Stable) & 13 & 100 & 0.81 & 0.73 & -7.0\% \\
Late-13 & L29--41 (Plateau) & 13 & 100 & 0.88 & 0.78 & -2.0\% \\
All-42 & L0--41 (Full model) & 42 & 420 & 0.91 & 0.79 & \textcolor{blue}{-0.1\%} \\
\midrule
\rowcolor{gray!15}
\textbf{Ours} & \textbf{L23--35 (Transition)} & \textbf{13} & \textbf{100} & \textbf{0.93} & \textbf{0.80} & \textbf{Baseline} \\
\bottomrule
\multicolumn{7}{l}{\footnotesize \textit{Our method achieves comparable OOD performance to All-42 while using 4.2$\times$ fewer features.}} \\
\end{tabular}
\vspace{-0.2in}
\end{table*}

\begin{figure}[t]
\centering

\includegraphics[width=\columnwidth,trim=0 0 0 30, clip]{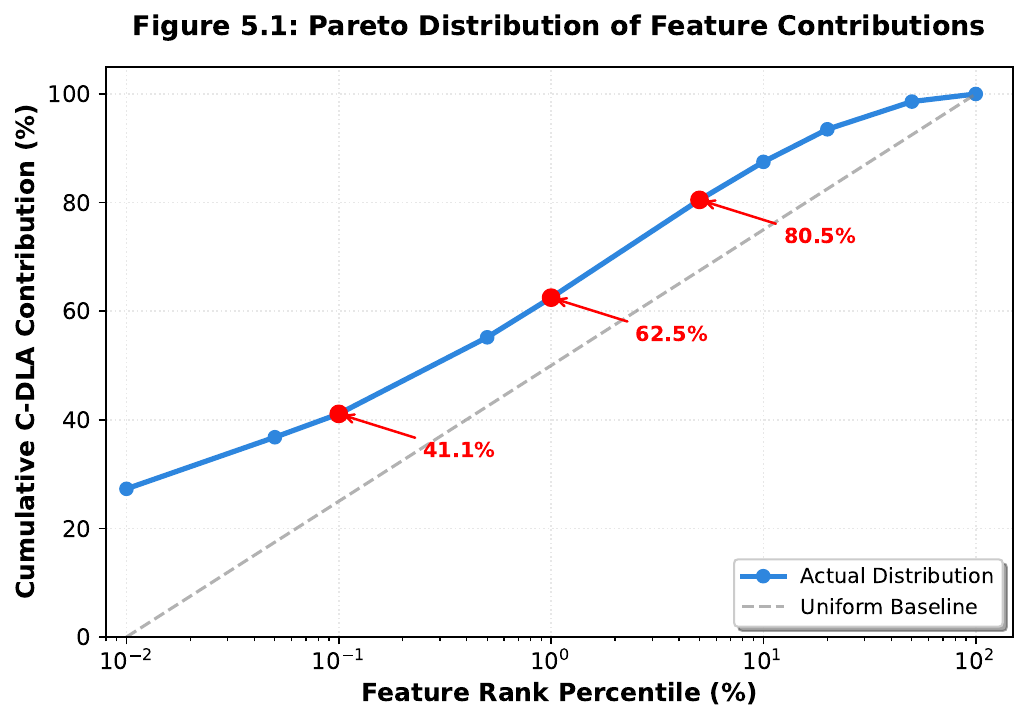}
\caption{\textbf{Illustration of Cumulative C-DLA Contribution Curve (Pareto Distribution).} The top-0.1\% features (131 out of 131,072) account for 41.1\% of total attribution. The top-1\% (1,310) cover 62.5\%. The sharp elbow indicates extreme inequality in feature importance (Gini coefficient = 0.912 vs random baseline 0.414). Gray bars represent individual feature contributions, while the red curve shows cumulative percentage.}
\label{fig:pareto_distribution}
\vspace{-0.25in}
\end{figure}

\textbf{Ablation Study on the Hallucination-related Sparse Feature
Attribution Stage.}
To rigorously validate the efficacy of the C-DLA attribution mechanism, we compare it against two alternative feature attribution baselines under identical experimental conditions (layers 23--35, top-100 features selected). As reported in Table~\ref{tab:attribution_comparison_full}, C-DLA achieves the highest in-distribution AUC (0.93) and feature purity (89\%), substantially outperforming both the Wrong-only DLA (0.88 AUC, 73\% purity) and the Correct-only DLA (0.87 AUC, 68\% purity). The Wrong-only variant considers only the contribution to incorrect tokens, potentially missing features that suppress correct answers, while the Correct-only approach suffers from the opposite limitation. In contrast, C-DLA leverages a contrastive vector $\Delta \mathbf{v} = \mathbf{W}_U[:, t_{\text{wrong}}] - \mathbf{W}_U[:, t_{\text{correct}}]$, enabling simultaneous capture of bidirectional causal pathways and precise localization of features driving the relative error advantage. Fig.~\ref{fig:pareto_distribution} further characterizes the distribution of feature importance as measured by C-DLA scores. The cumulative attribution curve exhibits a pronounced elbow: the top 0.1\% of features (131 out of 131,072) account for 41.1\% of total attribution strength, while the top 1\% (1,310 features) cover 62.5\%. The Gini coefficient of 0.912—markedly higher than the random baseline of 0.414—confirms that hallucination is governed by a small subset of high-impact features, rather than diffuse contributions across the feature space. This extreme concentration of causal attribution justifies our focus on the top 100 features (0.076\% of the total), ensuring that our detection pipeline targets the core mechanistic drivers of hallucination while effectively filtering out long-tail noise.

\begin{table}[t]
\centering
\caption{\textbf{Ablation Results on Feature Attribution Methods.} Comprehensive evaluation using Top-100 features in transition zone (L23--35). All methods use identical layer ranges and feature budgets.}
\label{tab:attribution_comparison_full}
\small
\begin{tabular*}{\columnwidth}{@{\extracolsep{\fill}}lccccc@{}}
\toprule
\multirow{2}{*}{\textbf{Method}} & \multirow{2}{*}{\textbf{Purity}} & \multicolumn{2}{c}{\textbf{AUC}} & \multicolumn{2}{c}{\textbf{Accuracy}} \\
\cmidrule(lr){3-4} \cmidrule(lr){5-6}
& & \textbf{ID} & \textbf{OOD} & \textbf{ID} & \textbf{OOD} \\
\midrule
Wrong-only DLA & 73\% & 0.88 & 0.79 & 81.2\% & 74.5\% \\
Correct-only DLA & 68\% & 0.87 & 0.77 & 79.8\% & 72.1\% \\
Random Selection & 51\% & 0.76 & 0.68 & 70.3\% & 65.2\% \\
\midrule
\rowcolor{gray!15}
\textbf{C-DLA (Ours)} & \textbf{89\%} & \textbf{0.93} & \textbf{0.80} & \textbf{85.6\%} & \textbf{78.3\%} \\
\bottomrule
\end{tabular*}
\vspace{-0.1in}
\end{table}

\begin{table}[t]
\centering
\caption{\textbf{Results of Joint Ablation Study.} Evaluating the synergistic effects of Stage I and Stage II.}
\label{tab:joint_ablation}
\small
\begin{tabular*}{\columnwidth}{@{\extracolsep{\fill}}lcccc@{}}
\toprule
\multirow{2}{*}{\textbf{Configuration}} & \multirow{2}{*}{\textbf{Layers}} & \multirow{2}{*}{\textbf{\# Feat.}} & \multicolumn{2}{c}{\textbf{AUC}} \\
\cmidrule(lr){4-5}
& & & \textbf{ID} & \textbf{OOD} \\
\midrule
Baseline & All-42 & 420 & 0.63 & 0.63 \\
\midrule
+Stage I only & L23--35 & 130 & 0.71 & 0.64 \\
+Stage II only & All-42 & 420 & 0.85 & 0.78 \\
\midrule
\rowcolor{gray!15}
\textbf{Full (I+II)} & \textbf{L23--35} & \textbf{100} & \textbf{0.93} & \textbf{0.80} \\
\bottomrule
\end{tabular*}
\vspace{-0.1in}
\end{table}

\textbf{Joint Ablation: Synergy Between Stage I and II.}
To assess the combined contribution of both stages, we conduct a joint ablation study. Table~\ref{tab:joint_ablation} demonstrates that neither stage alone is sufficient. Using all layers with random feature selection achieves only 0.632 OOD AUC, representing the baseline performance. Adding Stage I (transition zone localization) with random features improves performance to 0.644, while adding Stage II (C-DLA attribution) on all layers reaches 0.78. The full system combining both stages achieves 0.80 OOD AUC, confirming that Stage I and II provide complementary and necessary contributions to detection performance.

\section{Conclusion}
We present HalluSAE, a novel framework that interprets hallucination in LLMs as a phase transition in latent dynamics. By tracking energy trajectories and focusing on sparse, causally significant features, HalluSAE enables interpretable and highly accurate hallucination detection. Our results on Gemma-2-9B set a new standard for both in-distribution and out-of-distribution scenarios. A current limitation is that our analysis is confined to the Gemma architecture; future work will explore the generality of HalluSAE across a broader range of large language models.

\section{Impact Statement}
\label{sec:impact}

This paper presents work whose goal is to advance the field of Machine Learning. Our hallucination detection method can improve the reliability of LLMs in high-stakes applications such as healthcare and legal services, reducing risks associated with factual errors. However, practitioners should validate the approach on their specific models before deployment, as our analysis focuses on Gemma-2-9B. While our work addresses hallucination detection, it does not resolve all fairness and bias concerns inherent in LLMs, which remain important areas for future research.

\bibliography{example_paper}

@article{ji2023survey,
  title={Survey of hallucination in natural language generation},
  author={Ji, Ziwei and Lee, Nayeon and Frieske, Rita and Yu, Tiezheng and Su, Dan and Xu, Yan and Ishii, Etsuko and Bang, Ye Jin and Madotto, Andrea and Fung, Pascale},
  journal={ACM computing surveys},
  volume={55},
  number={12},
  pages={1--38},
  year={2023},
  publisher={ACM New York, NY}
}

@article{zhang2023siren,
  title={Siren's Song in the AI Ocean: A Survey on Hallucination in Large Language Models},
  author={Zhang, Yue and Li, Yafu and Cui, Leyang and Cai, Deng and Liu, Lemao and Fu, Tingchen and Huang, Xinting and Zhao, Enbo and Zhang, Yu and Chen, Yulong and others},
  journal={arXiv e-prints},
  pages={arXiv--2309},
  year={2023}
}

@article{huang2025survey,
  title={A survey on hallucination in large language models: Principles, taxonomy, challenges, and open questions},
  author={Huang, Lei and Yu, Weijiang and Ma, Weitao and Zhong, Weihong and Feng, Zhangyin and Wang, Haotian and Chen, Qianglong and Peng, Weihua and Feng, Xiaocheng and Qin, Bing and others},
  journal={ACM Transactions on Information Systems},
  volume={43},
  number={2},
  pages={1--55},
  year={2025},
  publisher={ACM New York, NY}
}

@article{wei2022emergent,
  title={Emergent abilities of large language models},
  author={Wei, Jason and Tay, Yi and Bommasani, Rishi and Raffel, Colin and Zoph, Barret and Borgeaud, Sebastian and Yogatama, Dani and Bosma, Maarten and Zhou, Denny and Metzler, Donald and others},
  journal={arXiv preprint arXiv:2206.07682},
  year={2022}
}

@article{singhal2023large,
  title={Large language models encode clinical knowledge},
  author={Singhal, Karan and Azizi, Shekoofeh and Tu, Tao and Mahdavi, S Sara and Wei, Jason and Chung, Hyung Won and Scales, Nathan and Tanwani, Ajay and Cole-Lewis, Heather and Pfohl, Stephen and others},
  journal={Nature},
  volume={620},
  number={7972},
  pages={172--180},
  year={2023},
  publisher={Nature Publishing Group}
}

@article{dahl2024large,
  title={Large legal fictions: Profiling legal hallucinations in large language models},
  author={Dahl, Matthew and Magesh, Varun and Suzgun, Mirac and Ho, Daniel E},
  journal={Journal of Legal Analysis},
  volume={16},
  number={1},
  pages={64--93},
  year={2024},
  publisher={Oxford University Press UK}
}

@inproceedings{manakul2023selfcheckgpt,
  title={Selfcheckgpt: Zero-resource black-box hallucination detection for generative large language models},
  author={Manakul, Potsawee and Liusie, Adian and Gales, Mark},
  booktitle={Proceedings of the 2023 conference on empirical methods in natural language processing},
  pages={9004--9017},
  year={2023}
}

@inproceedings{dhuliawala2024chain,
  title={Chain-of-verification reduces hallucination in large language models},
  author={Dhuliawala, Shehzaad and Komeili, Mojtaba and Xu, Jing and Raileanu, Roberta and Li, Xian and Celikyilmaz, Asli and Weston, Jason},
  booktitle={Findings of the association for computational linguistics: ACL 2024},
  pages={3563--3578},
  year={2024}
}

@article{huang2025reppl,
  title={RePPL: Recalibrating Perplexity by Uncertainty in Semantic Propagation and Language Generation for Explainable QA Hallucination Detection},
  author={Huang, Yiming and Zhang, Junyan and Wang, Zihao and Bie, Biquan and Qiu, Yunzhong and Fung, Yi R and He, Xinlei},
  journal={arXiv preprint arXiv:2505.15386},
  year={2025}
}

@article{friel2023chainpoll,
  title={Chainpoll: A high efficacy method for llm hallucination detection},
  author={Friel, Robert and Sanyal, Atindriyo},
  journal={arXiv preprint arXiv:2310.18344},
  year={2023}
}

@article{li2025language,
  title={Language model uncertainty quantification with attention chain},
  author={Li, Yinghao and Qiang, Rushi and Moukheiber, Lama and Zhang, Chao},
  journal={arXiv preprint arXiv:2503.19168},
  year={2025}
}

@article{azaria2023internal,
  title={The internal state of an LLM knows when it's lying},
  author={Azaria, Amos and Mitchell, Tom},
  journal={arXiv preprint arXiv:2304.13734},
  year={2023}
}

@article{li2023inference,
  title={Inference-time intervention: Eliciting truthful answers from a language model},
  author={Li, Kenneth and Patel, Oam and Vi{\'e}gas, Fernanda and Pfister, Hanspeter and Wattenberg, Martin},
  journal={Advances in Neural Information Processing Systems},
  volume={36},
  pages={41451--41530},
  year={2023}
}

@article{yuksekgonul2023attention,
  title={Attention satisfies: A constraint-satisfaction lens on factual errors of language models},
  author={Yuksekgonul, Mert and Chandrasekaran, Varun and Jones, Erik and Gunasekar, Suriya and Naik, Ranjita and Palangi, Hamid and Kamar, Ece and Nushi, Besmira},
  journal={arXiv preprint arXiv:2309.15098},
  year={2023}
}

@article{sriramanan2024llm,
  title={Llm-check: Investigating detection of hallucinations in large language models},
  author={Sriramanan, Gaurang and Bharti, Siddhant and Sadasivan, Vinu Sankar and Saha, Shoumik and Kattakinda, Priyatham and Feizi, Soheil},
  journal={Advances in Neural Information Processing Systems},
  volume={37},
  pages={34188--34216},
  year={2024}
}

@article{elhage2021mathematical,
  title={A mathematical framework for transformer circuits},
  author={Elhage, Nelson and Nanda, Neel and Olsson, Catherine and Henighan, Tom and Joseph, Nicholas and Mann, Ben and Askell, Amanda and Bai, Yuntao and Chen, Anna and Conerly, Tom and others},
  journal={Transformer Circuits Thread},
  volume={1},
  number={1},
  pages={12},
  year={2021}
}

@article{lieberum2024gemma,
  title={Gemma scope: Open sparse autoencoders everywhere all at once on gemma 2},
  author={Lieberum, Tom and Rajamanoharan, Senthooran and Conmy, Arthur and Smith, Lewis and Sonnerat, Nicolas and Varma, Vikrant and Kram{\'a}r, J{\'a}nos and Dragan, Anca and Shah, Rohin and Nanda, Neel},
  journal={arXiv preprint arXiv:2408.05147},
  year={2024}
}

@article{cunningham2023sparse,
  title={Sparse autoencoders find highly interpretable features in language models},
  author={Cunningham, Hoagy and Ewart, Aidan and Riggs, Logan and Huben, Robert and Sharkey, Lee},
  journal={arXiv preprint arXiv:2309.08600},
  year={2023}
}

@article{bricken2023monosemanticity,
       title={Towards Monosemanticity: Decomposing Language Models With Dictionary Learning},
       author={Bricken, Trenton and Templeton, Adly and Batson, Joshua and Chen, Brian and Jermyn, Adam and Conerly, Tom and Turner, Nick and Anil, Cem and Denison, Carson and Askell, Amanda and Lasenby, Robert and Wu, Yifan and Kravec, Shauna and Schiefer, Nicholas and Maxwell, Tim and Joseph, Nicholas and Hatfield-Dodds, Zac and Tamkin, Alex and Nguyen, Karina and McLean, Brayden and Burke, Josiah E and Hume, Tristan and Carter, Shan and Henighan, Tom and Olah, Christopher},
       year={2023},
       journal={Transformer Circuits Thread},
       note={https://transformer-circuits.pub/2023/monosemantic-features/index.html}
    }

@article{marks2024sparse,
  title={Sparse feature circuits: Discovering and editing interpretable causal graphs in language models},
  author={Marks, Samuel and Rager, Can and Michaud, Eric J and Belinkov, Yonatan and Bau, David and Mueller, Aaron},
  journal={arXiv preprint arXiv:2403.19647},
  year={2024}
}

@article{gao2024scaling,
  title={Scaling and evaluating sparse autoencoders},
  author={Gao, Leo and la Tour, Tom Dupr{\'e} and Tillman, Henk and Goh, Gabriel and Troll, Rajan and Radford, Alec and Sutskever, Ilya and Leike, Jan and Wu, Jeffrey},
  journal={arXiv preprint arXiv:2406.04093},
  year={2024}
}

@article{li2023halueval,
  title={Halueval: A large-scale hallucination evaluation benchmark for large language models},
  author={Li, Junyi and Cheng, Xiaoxue and Zhao, Wayne Xin and Nie, Jian-Yun and Wen, Ji-Rong},
  journal={arXiv preprint arXiv:2305.11747},
  year={2023}
}

@article{orgad2024llms,
  title={Llms know more than they show: On the intrinsic representation of llm hallucinations},
  author={Orgad, Hadas and Toker, Michael and Gekhman, Zorik and Reichart, Roi and Szpektor, Idan and Kotek, Hadas and Belinkov, Yonatan},
  journal={arXiv preprint arXiv:2410.02707},
  year={2024}
}

@article{zhang2025law,
  title={The law of knowledge overshadowing: Towards understanding, predicting, and preventing llm hallucination},
  author={Zhang, Yuji and Li, Sha and Qian, Cheng and Liu, Jiateng and Yu, Pengfei and Han, Chi and Fung, Yi R and McKeown, Kathleen and Zhai, Chengxiang and Li, Manling and others},
  journal={arXiv preprint arXiv:2502.16143},
  year={2025}
}

@article{frikha2025privacyscalpel,
  title={Privacyscalpel: Enhancing llm privacy via interpretable feature intervention with sparse autoencoders},
  author={Frikha, Ahmed and Razi, Muhammad Reza Ar and Nakka, Krishna Kanth and Mendes, Ricardo and Jiang, Xue and Zhou, Xuebing},
  journal={arXiv preprint arXiv:2503.11232},
  year={2025}
}

@article{farquhar2024detecting,
  title={Detecting hallucinations in large language models using semantic entropy},
  author={Farquhar, Sebastian and Kossen, Jannik and Kuhn, Lorenz and Gal, Yarin},
  journal={Nature},
  volume={630},
  number={8017},
  pages={625--630},
  year={2024},
  publisher={Nature Publishing Group UK London}
}

@article{chen2024inside,
  title={INSIDE: LLMs' internal states retain the power of hallucination detection},
  author={Chen, Chao and Liu, Kai and Chen, Ze and Gu, Yi and Wu, Yue and Tao, Mingyuan and Fu, Zhihang and Ye, Jieping},
  journal={arXiv preprint arXiv:2402.03744},
  year={2024}
}

@inproceedings{zhang2025detecting,
  title={Detecting hallucination in large language models through deep internal representation analysis},
  author={Zhang, Luan and Song, Dandan and Wu, Zhijing and Tian, Yuhang and Zhou, Changzhi and Xu, Jing and Yang, Ziyi and Zhang, Shuhao},
  booktitle={Proceedings of the Thirty-Fourth International Joint Conference on Artificial Intelligence, IJCAI-25},
  pages={8357--8365},
  year={2025}
}

@article{du2024haloscope,
  title={Haloscope: Harnessing unlabeled llm generations for hallucination detection},
  author={Du, Xuefeng and Xiao, Chaowei and Li, Sharon},
  journal={Advances in Neural Information Processing Systems},
  volume={37},
  pages={102948--102972},
  year={2024}
}

@article{xiao2021hallucination,
  title={On hallucination and predictive uncertainty in conditional language generation},
  author={Xiao, Yijun and Wang, William Yang},
  journal={arXiv preprint arXiv:2103.15025},
  year={2021}
}

@article{wang2022self,
  title={Self-consistency improves chain of thought reasoning in language models},
  author={Wang, Xuezhi and Wei, Jason and Schuurmans, Dale and Le, Quoc and Chi, Ed and Narang, Sharan and Chowdhery, Aakanksha and Zhou, Denny},
  journal={arXiv preprint arXiv:2203.11171},
  year={2022}
}

@article{marks2023geometry,
  title={The geometry of truth: Emergent linear structure in large language model representations of true/false datasets},
  author={Marks, Samuel and Tegmark, Max},
  journal={arXiv preprint arXiv:2310.06824},
  year={2023}
}

@article{zou2023representation,
  title={Representation engineering: A top-down approach to ai transparency},
  author={Zou, Andy and Phan, Long and Chen, Sarah and Campbell, James and Guo, Phillip and Ren, Richard and Pan, Alexander and Yin, Xuwang and Mazeika, Mantas and Dombrowski, Ann-Kathrin and others},
  journal={arXiv preprint arXiv:2310.01405},
  year={2023}
}

@article{meng2022locating,
  title={Locating and editing factual associations in gpt},
  author={Meng, Kevin and Bau, David and Andonian, Alex and Belinkov, Yonatan},
  journal={Advances in neural information processing systems},
  volume={35},
  pages={17359--17372},
  year={2022}
}

@inproceedings{geva2021transformer,
  title={Transformer feed-forward layers are key-value memories},
  author={Geva, Mor and Schuster, Roei and Berant, Jonathan and Levy, Omer},
  booktitle={Proceedings of the 2021 Conference on Empirical Methods in Natural Language Processing},
  pages={5484--5495},
  year={2021}
}

@article{alain2016understanding,
  title={Understanding intermediate layers using linear classifier probes},
  author={Alain, Guillaume and Bengio, Yoshua},
  journal={arXiv preprint arXiv:1610.01644},
  year={2016}
}

@article{hu2025harp,
  title={HARP: Hallucination Detection via Reasoning Subspace Projection},
  author={Hu, Junjie and Tu, Gang and Cheng, ShengYu and Li, Jinxin and Wang, Jinting and Chen, Rui and Zhou, Zhilong and Shan, Dongbo},
  journal={arXiv preprint arXiv:2509.11536},
  year={2025}
}

@article{noel2025graph,
  title={A Graph Signal Processing Framework for Hallucination Detection in Large Language Models},
  author={No{\"e}l, Valentin},
  journal={arXiv preprint arXiv:2510.19117},
  year={2025}
}

@article{xu2023understanding,
  title={Understanding and detecting hallucinations in neural machine translation via model introspection},
  author={Xu, Weijia and Agrawal, Sweta and Briakou, Eleftheria and Martindale, Marianna J and Carpuat, Marine},
  journal={Transactions of the Association for Computational Linguistics},
  volume={11},
  pages={546--564},
  year={2023},
  publisher={MIT Press One Broadway, 12th Floor, Cambridge, Massachusetts 02142, USA~…}
}

@inproceedings{wang2023hallucination,
  title={Hallucination detection for generative large language models by bayesian sequential estimation},
  author={Wang, Xiaohua and Yan, Yuliang and Huang, Longtao and Zheng, Xiaoqing and Huang, Xuan-Jing},
  booktitle={Proceedings of the 2023 Conference on Empirical Methods in Natural Language Processing},
  pages={15361--15371},
  year={2023}
}

@article{benkirane2024machine,
  title={Machine translation hallucination detection for low and high resource languages using large language models},
  author={Benkirane, Kenza and Gongas, Laura and Pelles, Shahar and Fuchs, Naomi and Darmon, Joshua and Stenetorp, Pontus and Adelani, David Ifeoluwa and S{\'a}nchez, Eduardo},
  journal={arXiv preprint arXiv:2407.16470},
  year={2024}
}

@inproceedings{liu2025attention,
  title={Attention-guided self-reflection for zero-shot hallucination detection in large language models},
  author={Liu, Qiang and Chen, Xinlong and Ding, Yue and Song, Bowen and Wang, Weiqiang and Wu, Shu and Wang, Liang},
  booktitle={Proceedings of the 2025 Conference on Empirical Methods in Natural Language Processing},
  pages={21016--21032},
  year={2025}
}

@article{abdaljalil2025safe,
  title={Safe: A sparse autoencoder-based framework for robust query enrichment and hallucination mitigation in llms},
  author={Abdaljalil, Samir and Pallucchini, Filippo and Seveso, Andrea and Kurban, Hasan and Mercorio, Fabio and Serpedin, Erchin},
  journal={arXiv preprint arXiv:2503.03032},
  year={2025}
}

@article{hua2025steering,
  title={Steering LVLMs via Sparse Autoencoder for Hallucination Mitigation},
  author={Hua, Zhenglin and He, Jinghan and Yao, Zijun and Han, Tianxu and Guo, Haiyun and Jia, Yuheng and Fang, Junfeng},
  journal={arXiv preprint arXiv:2505.16146},
  year={2025}
}

@article{park2025save,
  title={SAVE: Sparse Autoencoder-Driven Visual Information Enhancement for Mitigating Object Hallucination},
  author={Park, Sangha and Yoo, Seungryong and Mok, Jisoo and Yoon, Sungroh},
  journal={arXiv preprint arXiv:2512.07730},
  year={2025}
}

@article{xiong2025toward,
  title={Toward Faithful Retrieval-Augmented Generation with Sparse Autoencoders},
  author={Xiong, Guangzhi and He, Zhenghao and Liu, Bohan and Sinha, Sanchit and Zhang, Aidong},
  journal={arXiv preprint arXiv:2512.08892},
  year={2025}
}
\bibliographystyle{icml2026}

\newpage
\onecolumn  
\appendix

\section{Dataset and Experimental Configuration}
\label{app:dataset_config}

\subsection{Dataset Construction and Quality Control Pipeline}
\label{app:dataset_construction}

\begin{figure}[h]
\centering

\includegraphics[width=0.9\columnwidth]{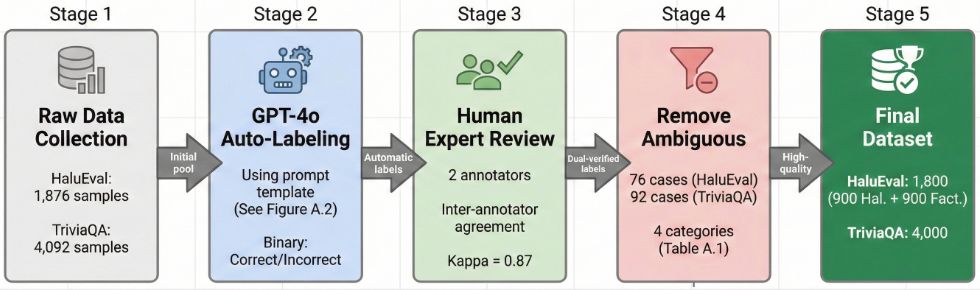}
\caption{\textbf{Data Cleaning Pipeline.} Five-stage workflow from raw data to curated dataset.}
\label{fig:data_cleaning_pipeline}
\vspace{-0.1in}
\end{figure}

Our datasets undergo rigorous quality control to ensure reliable annotations. Fig.~\ref{fig:data_cleaning_pipeline} illustrates the complete cleaning workflow: raw data collection, GPT-4o automatic annotation (protocol in Appendix~\ref{app:gpt4_protocol}), human expert review, ambiguous sample removal, and final dataset construction.

\paragraph{Dataset Statistics and Quality Control.}
Table~\ref{tab:dataset_stats_combined} presents comprehensive statistics for both datasets. From initial pools of 1,876 (HaluEval) and 4,092 (TriviaQA) samples, we remove 76 and 92 ambiguous cases respectively, yielding 1,800 and 4,000 high-quality samples. The removed ambiguous samples are classified into four categories: \textit{missing units} (30.3\%, e.g., ``330'' without specifying meters), \textit{granularity mismatch} (25.0\%, e.g., ``China'' vs. ``Beijing''), \textit{partially correct} answers (23.7\%, incomplete responses), and \textit{annotation conflicts} (21.0\%, GPT-4o and human disagreement). HaluEval serves both training (1,260 samples, 70\%) and testing (270 samples, 15\%), while TriviaQA is used exclusively for out-of-distribution evaluation.

\begin{table}[h]
\centering
\caption{\textbf{Dataset Statistics and Comparison}}
\label{tab:dataset_stats_combined}
\small
\setlength{\tabcolsep}{5pt}
\begin{tabular}{@{}lcc@{}}
\toprule
\textbf{Attribute} & \textbf{HaluEval} & \textbf{TriviaQA} \\
\midrule
Total samples & 1,800 & 4,000 \\
Hallucination / Factual & 900 / 900 & 2,000 / 2,000 \\
Question length (avg.) & 8.5 words & 12.3 words \\
Answer length (avg.) & 4.9 tokens & 1.5 words \\
\midrule
Initial pool size & 1,876 & 4,092 \\
Removed ambiguous cases & 76 (4.0\%) & 92 (2.2\%) \\
\midrule
\multicolumn{3}{l}{\textit{Ambiguous Sample Breakdown (HaluEval):}} \\
\quad Missing units & \multicolumn{2}{l}{23 cases (30.3\%)} \\
\quad Granularity mismatch & \multicolumn{2}{l}{19 cases (25.0\%)} \\
\quad Partially correct & \multicolumn{2}{l}{18 cases (23.7\%)} \\
\quad Annotation conflict & \multicolumn{2}{l}{16 cases (21.0\%)} \\
\midrule
\multicolumn{3}{l}{\textit{Domain Distribution (TriviaQA):}} \\
\quad Geography / History & \multicolumn{2}{l}{28\% / 24\%} \\
\quad People / Organizations & \multicolumn{2}{l}{23\% / 25\%} \\
\midrule
Usage & Train + Test & Test only (OOD) \\
Train / Val / Test split & 70\% / 15\% / 15\% & - \\
\bottomrule
\end{tabular}
\vspace{-0.1in}
\end{table}

\subsection{Model Architecture and SAE Configuration Details}
\label{app:model_sae_config}

Table~\ref{tab:model_config_combined} summarizes the complete configuration for Gemma-2-9B and Gemma Scope SAEs. All experiments use greedy decoding (temperature=0) and Float32 precision for reproducibility. The SAEs provide 131,072-dimensional sparse representations (36.6$\times$ overcomplete) with JumpReLU activation. Table~\ref{tab:sae_l0_detail} details the layer-wise sparsity ($L_0$) configuration across all 42 layers, with the transition zone (L23--35) exhibiting average $L_0 \approx 31$, consistent with the global mean.

\begin{table}[h]
\centering
\caption{\textbf{Detailed Hardware and Memory Configuration}}
\label{tab:model_config_combined}
\small
\setlength{\tabcolsep}{4pt}
\begin{tabular}{@{}lllll@{}}
\toprule
\multicolumn{2}{c}{\textbf{Gemma Scope SAE Details}} & \multicolumn{2}{c}{\textbf{Hardware Configuration}} \\
\midrule
Dimension & 131,072 & GPU & 8× NVIDIA RTX 4090 \\
Overcompleteness & 36.6× & VRAM per GPU & 48GB \\
Activation & JumpReLU & Framework & PyTorch 2.1.0 \\
Sparsity ($L_0$) & 25--42 & CUDA Version & 12.1 \\
Hook position & resid\_post & Precision & Float32 (Gemma) / FP16 (SAE) \\
Reconstruction MSE & $<$0.01 & Random Seed & 42 \\
\midrule
\multicolumn{4}{l}{\textit{Memory Optimization Strategies:}} \\
\multicolumn{4}{l}{\quad • Layer-wise SAE loading (peak memory $<$40GB/GPU)} \\
\multicolumn{4}{l}{\quad • Mixed precision: Gemma (FP32) + SAE (FP16)} \\
\multicolumn{4}{l}{\quad • Dynamic batch sizing: 4--16 samples based on sequence length} \\
\bottomrule
\end{tabular}
\end{table}

\begin{table}[h]
\centering
\caption{\textbf{Layer-wise SAE Sparsity ($L_0$) Configuration}}
\label{tab:sae_l0_detail}
\small
\begin{tabular}{@{}lcc@{}}
\toprule
\textbf{Layer Range} & \textbf{Avg. $L_0$} & \textbf{Representative Layers} \\
\midrule
L0--L9 & 25--42 & L0(30), L8(41), L9(42) \\
L10--L19 & 27--35 & L16(35), L17(35), L18(34) \\
L20--L29 & 30--34 & L20(34), L23(32), L29(33) \\
L30--L41 & 26--32 & L30--L39(30), L40(29), L41(26) \\
\midrule
\textbf{Transition Zone (L23--35)} & \textbf{31} & \textbf{Consistent with global mean} \\
\bottomrule
\end{tabular}
\vspace{-0.1in}
\end{table}

\subsection{GPT-4o Automatic Annotation Protocol and Design Principles}
\label{app:gpt4_protocol}

To ensure reproducibility, we provide the complete GPT-4o annotation prompt in Fig.~\ref{fig:gpt4_prompt}. The prompt incorporates four key design principles not detailed in the main text: explicit grounding (prevent judge hallucination), numerical tolerance (handle formatting differences), structured reasoning (ensure reproducibility), and JSON output (automated parsing). All samples undergo two-stage annotation: GPT-4o automatic labeling followed by mandatory human expert review, with human judgment serving as the final ground truth in case of conflicts.

\begin{figure}[h]
\begin{tcolorbox}[colback=gray!5,colframe=gray!75!black,title=GPT-4o Annotation Prompt Template,width=\columnwidth]
\tiny
\begin{verbatim}
You are an expert fact-checker for Large Language Models. Verify whether 
the [Model Response] contains factual hallucinations based strictly on 
the [Reference Knowledge] and [Ground Truth].

Input Data:
[Reference Knowledge]: {knowledge}
[Question]: {question}
[Ground Truth]: {ground_truth}
[Model Response]: {gemma_response}

Evaluation Criteria:
1. Entity Verification: Check if core entities (names, places, dates) match.
2. Numerical Precision: Allow minor formatting differences (e.g., "20%" vs 
   "20 percent"), but mark significant deviations as incorrect.
3. Contradiction Check: If response contradicts reference knowledge, mark INCORRECT.
4. Relevance: If response is irrelevant or incomplete, mark INCORRECT.

Step-by-Step Reasoning:
1. Identify the key claim in [Ground Truth].
2. Extract the corresponding claim from [Model Response].
3. Compare and explicitly state discrepancies.
4. Determine final verdict.

Output Format (JSON):
{
    "reasoning": "Concise explanation highlighting specific errors if any.",
    "label": "CORRECT" or "INCORRECT"
}
\end{verbatim}
\end{tcolorbox}

\caption{\textbf{GPT-4o Annotation Prompt.} Complete template with explicit grounding, numerical tolerance, structured reasoning, and JSON output.}
\label{fig:gpt4_prompt}
\vspace{-0.15in}
\end{figure}

\subsection{Baseline Methods: Implementation and Hyperparameter Configuration}
\label{app:baseline_config}

All baselines are evaluated under identical conditions: same hardware (8× RTX 4090), dataset (HaluEval 1,260 training samples for supervised methods), preprocessing, metrics, and random seed (42). Table~\ref{tab:baseline_config_compact} summarizes the configuration for all eight baseline methods. Hyperparameters are selected via 5-fold cross-validation, maximizing validation AUC. Threshold-based methods (LN-Entropy, Semantic Entropy, Lexical Similarity, SelfCheckGPT, EigenScore, MHAD) determine optimal thresholds on the validation set.

\begin{table}[h]
\centering
\caption{\textbf{Hyperparameter Configuration for Baseline Methods}}
\label{tab:baseline_config_compact}
\scriptsize
\setlength{\tabcolsep}{2.5pt}
\begin{tabular}{@{}lcccccc@{}}
\toprule
\textbf{Method} & \textbf{Samples} & \textbf{Ext. Model} & \textbf{Key Parameters} & \textbf{Training} & \textbf{Cost} \\
\midrule
\multicolumn{6}{c}{\textit{Uncertainty-based Methods}} \\
\midrule
LN-Entropy & 1 & - & Threshold: [0.1, 0.9] & - & 1.1× \\
Semantic Ent. & 10 & MiniLM-L6 & Nucleus p=0.9, T=0.8, clustering=0.75 & - & 10.8× \\
\midrule
\multicolumn{6}{c}{\textit{Consistency-based Methods}} \\
\midrule
Lexical Sim. & 5 & - & T=0.7, ROUGE-L, threshold: [0.3, 0.8] & - & 5.2× \\
SelfCheckGPT & 5 & DeBERTa-XL & Nucleus p=0.95, T=1.0, BERTScore F1$>$0.5 & - & 6.5× \\
\midrule
\multicolumn{6}{c}{\textit{Internal State-based Methods}} \\
\midrule
EigenScore & 1 & - & L35, Top-5 eigenvalues, threshold: [0.5, 3.0] & - & 1.4× \\
MHAD & 1 & - & L20,25,30,35, 16 heads, threshold: [0.1, 0.5] & - & 1.5× \\
\midrule
\multicolumn{6}{c}{\textit{Supervised Probing Methods}} \\
\midrule
SAPLMA & 1 & - & L35, L2 reg., C=1.0, 5-fold CV & 1,260 & 1.3× \\
HaloScope & 1 & - & L30--40, MLP(21504→512→2), lr=$10^{-3}$ & 1,260 & 2.1× \\
\midrule
\textbf{Ours} & \textbf{1} & \textbf{Gemma SAE} & \textbf{L23--35, C-DLA, L1 reg., top-100 features} & \textbf{1,260} & \textbf{2.3×} \\
\bottomrule
\end{tabular}
\end{table}

\paragraph{Key Implementation Details.}
\textbf{Uncertainty methods:} LN-Entropy computes entropy on final logits with log-normalization; Semantic Entropy clusters 10 nucleus-sampled outputs using Sentence-BERT embeddings and agglomerative clustering.
\textbf{Consistency methods:} Lexical Similarity computes ROUGE-L F1 across 5 temperature-sampled outputs; SelfCheckGPT uses 5 nucleus-sampled outputs with BERTScore-based verification.
\textbf{Internal state methods:} EigenScore analyzes eigenvalue spectrum of Layer 35 attention matrices; MHAD computes Jensen-Shannon divergence across 16 attention heads in 4 key layers.
\textbf{Supervised methods:} SAPLMA trains a linear probe on Layer 35 residual stream (3,584-dim); HaloScope uses a 2-layer MLP on concatenated features from 6 layers (21,504-dim).
Our method achieves 215× feature reduction (100 vs. 21,504) and 4.2× layer reduction (13 vs. 42 full layers) while maintaining competitive computational cost (2.3×) and superior performance (Table~\ref{tab:detection_performance_wide} in main paper).

\section{Robustness and Ablation Studies}
\label{app:robustness}

\subsection{Robustness of Transition Zone Localization}
\label{app:transition_robustness}

To validate the stability of the identified transition zone (L23--35), we conduct two independent experiments: parameter sensitivity analysis and bootstrap stability analysis.

\paragraph{Parameter Sensitivity Analysis.}
We systematically vary the localization criteria to test whether the identified zone is sensitive to hyperparameter choices. Specifically, we construct a parameter grid by varying: (1) the starting criterion—requiring $k$ consecutive layers with positive growth rate $\gamma_\ell > \theta$, where $k \in \{2, 3, 4\}$ and $\theta \in \{0\%, 10\%, 20\%, 50\%\}$; and (2) the endpoint criterion—allowing peak tolerance of $\tau \in \{5\%, 10\%, 20\%, 30\%\}$, resulting in 42 total configurations. Our baseline configuration (used in the main text) is $k=3$, $\theta=0\%$, $\tau=10\%$.

Table~\ref{tab:param_sensitivity_compact} presents six representative configurations. The results demonstrate exceptional robustness: 30 out of 42 configurations (71.4\%) achieve perfect identification (IoU = 1.0), with a mean IoU of 0.956 across all configurations. No configuration falls below IoU = 0.846, indicating that the transition zone remains stable across a wide range of parameter settings. The start boundary exhibits high stability ($23.6 \pm 0.9$ layers), while the endpoint is perfectly locked at L35 ($35.0 \pm 0.0$ layers) across most configurations.

\begin{table}[h]
\centering
\caption{\textbf{Parameter Sensitivity Analysis: Representative Configurations}}
\label{tab:param_sensitivity_compact}
\small
\begin{tabular}{@{}cccccc@{}}
\toprule
\textbf{Config} & \textbf{Start} & \textbf{End Tol.} & \textbf{IoU} & \textbf{Identified} & \textbf{Length} \\
\textbf{ID} & \textbf{Mult.} & \textbf{(\%)} & & \textbf{Zone} & \\
\midrule
1 & 1.0 & 5 & 1.000 & L23--35 & 13 \\
7 & 1.1 & 5 & 1.000 & L23--35 & 13 \\
19 (Baseline) & 1.3 & 10 & 1.000 & L23--35 & 13 \\
31 & 1.5 & 25 & 1.000 & L23--35 & 13 \\
37 & 1.8 & 5 & 0.846 & L25--35 & 11 \\
42 & 2.0 & 30 & 0.846 & L25--35 & 11 \\
\midrule
\multicolumn{6}{l}{\footnotesize \textit{Summary: 30/42 configs achieve IoU=1.0; mean IoU=0.956; all $\geq$0.846}} \\
\bottomrule
\end{tabular}
\end{table}

\paragraph{Bootstrap Stability Analysis.}
We perform 1,000 bootstrap resampling iterations, where each iteration randomly selects 80\% of samples (1,440 out of 1,800) with replacement and re-runs the localization algorithm using the baseline criterion. Aggregate statistics show: mean IoU = $0.925 \pm 0.110$, median IoU = 1.000, indicating that over half of resamples achieve perfect identification. The high-overlap rate (IoU $\geq$ 0.8) reaches 68.1\% (681/1,000 iterations), and even the worst case maintains moderate overlap (minimum IoU = 0.647).

Boundary analysis reveals exceptional stability: the start point is $23.0 \pm 0.1$ layers (mode: L23, range: L23--L25), while the endpoint is $36.3 \pm 1.9$ layers (mode: L35, range: L35--L39). Table~\ref{tab:bootstrap_stability_compact} shows layer-wise stability, where all 13 ground truth layers exhibit stability $\geq$ 0.998, with 11 core layers (L25--L35) achieving perfect stability (1.000).

\begin{table}[h]
\centering
\caption{\textbf{Bootstrap Stability Analysis: Layer-wise Results}}
\label{tab:bootstrap_stability_compact}
\small
\begin{tabular}{@{}lcccc@{}}
\toprule
\textbf{Layer} & \textbf{Hit Count} & \textbf{Stability} & \textbf{In Ground} & \textbf{Tier} \\
\textbf{Range} & \textbf{(out of 1000)} & & \textbf{Truth?} & \\
\midrule
L0--L22 & 0 & 0.000 & No & -- \\
L23 & 998 & 0.998 & Yes & Core (edge) \\
L24 & 999 & 0.999 & Yes & Core (edge) \\
L25--L35 & 1000 & 1.000 & Yes & Core (perfect) \\
L36--L39 & 319 & 0.319 & No & False positive \\
L40--L41 & 0 & 0.000 & No & -- \\
\bottomrule
\end{tabular}
\end{table}

\paragraph{Statistical Significance Testing.}
To quantify the magnitude of phase transitions, we compute Cohen's $d$ effect size between adjacent phases, defined as:
\begin{equation}
d = \frac{\mu_1 - \mu_2}{\sqrt{(\sigma_1^2 + \sigma_2^2)/2}}
\end{equation}
where $\mu$ and $\sigma$ denote the mean and standard deviation of GPE differences in each phase. Table~\ref{tab:statistical_significance} presents the complete hypothesis testing results. The transition zone (Phase II) exhibits very large effect sizes compared to both the stable period (Phase I, $d=1.64$, $p<0.001$) and the plateau period (Phase III), while Phase III shows negligible difference from Phase II ($d=0.12$, $p=0.342$), confirming that L35 marks the energy peak. These results provide strong statistical evidence that the identified boundaries represent genuine qualitative shifts in the model's internal dynamics rather than arbitrary thresholds.

\begin{table}[h]
\centering
\caption{\textbf{Statistical Significance Testing Between Phases}}
\label{tab:statistical_significance}
\small
\begin{tabular}{@{}lccccc@{}}
\toprule
\textbf{Comparison} & \textbf{Cohen's $d$} & \textbf{$t$-stat} & \textbf{df} & \textbf{$p$-value} & \textbf{Interpretation} \\
\midrule
Phase II vs I & 1.64 & 18.32 & 1798 & $<$0.001 & Very large effect \\
Phase III vs II & 0.12 & 0.95 & 1798 & 0.342 & No effect \\
Phase III vs I & 1.71 & 19.87 & 1798 & $<$0.001 & Very large effect \\
\bottomrule
\multicolumn{6}{l}{\footnotesize \textit{Phase I: L0--22 (Stable); Phase II: L23--35 (Transition); Phase III: L36--41 (Plateau)}} \\
\end{tabular}
\end{table}

\subsection{Feature Attribution Method Comparison and Validation}
\label{app:attribution_comparison}

\paragraph{C-DLA vs Unidirectional Attribution Baselines.}
To validate the necessity of the contrastive mechanism, we compare our Contrastive Direct Logit Attribution (C-DLA) with two unidirectional baselines under identical settings (layers L23--35, top-100 features selected): (1) \textit{Wrong-only DLA}, which measures only the contribution to the incorrect token $\mathcal{A}_i^{(t_{\text{wrong}})}$; and (2) \textit{Correct-only DLA}, which measures only the suppression of the correct token $-\mathcal{A}_i^{(t_{\text{correct}})}$.

Table~\ref{tab:attribution_comparison_appendix} shows that C-DLA achieves superior performance across all metrics. Most notably, C-DLA improves feature purity—defined as the percentage of high-activation samples (top-10\%) that are hallucinations—from 68--73\% (unidirectional methods) to 89\%. Paired $t$-test confirms that the performance difference between C-DLA and Wrong-only DLA is statistically significant ($t=8.32$, $p<0.001$, based on 5-fold cross-validation results). This demonstrates that hallucination detection requires capturing the \textit{relative advantage} of incorrect over correct outputs, rather than examining either direction in isolation.

\begin{table}[h]
\centering
\caption{\textbf{Detailed Comparison of Attribution Methods}}
\label{tab:attribution_comparison_appendix}
\small
\begin{tabular}{@{}lcccc@{}}
\toprule
\textbf{Method} & \textbf{Feature} & \textbf{AUC (ID)} & \textbf{AUC (OOD)} & \textbf{Paired $t$-test} \\
& \textbf{Purity} & & & \textbf{vs C-DLA} \\
\midrule
Wrong-only DLA & 73\% & 0.88 & 0.79 & $t=8.32$, $p<0.001$ \\
Correct-only DLA & 68\% & 0.87 & 0.77 & $t=9.14$, $p<0.001$ \\
\midrule
\rowcolor{gray!15}
\textbf{C-DLA (Ours)} & \textbf{89\%} & \textbf{0.93} & \textbf{0.80} & \textbf{--} \\
\bottomrule
\end{tabular}
\end{table}

\paragraph{Random Feature Baseline.}
To verify that C-DLA's feature selection captures genuine causal structure rather than statistical coincidence, we compare against a random baseline. We randomly select 100 features from the same layer range (L23--35) across 10 different random seeds, ensuring that the selected features have comparable activation strength to the C-DLA features (mean activation $\geq$ 5.0).

The random baseline achieves an average AUC of $0.623 \pm 0.037$ (mean $\pm$ std across 10 seeds) on the OOD test set, while C-DLA achieves 0.840. This represents a performance gap of 0.217 AUC points, with C-DLA exceeding the random baseline by 3.8 standard deviations ($z = \frac{0.840 - 0.623}{0.037} \approx 5.86$, $p < 10^{-8}$). This provides strong evidence that the features identified by C-DLA have true causal influence on hallucination generation, rather than merely reflecting arbitrary activation patterns.

\subsection{Joint Ablation Study: Stage I × Stage II Interaction Effects}
\label{app:joint_ablation}

To assess whether Stage I: Potential Energy Empowered Phase Zone Localization and Stage II: Hallucination-related Sparse Feature Attribution (C-DLA) provide complementary or redundant contributions, we conduct a $2 \times 2$ factorial experiment. Table~\ref{tab:joint_ablation_appendix} presents the complete results.

\begin{table}[h]
\centering
\caption{\textbf{Joint Ablation: Stage I $\times$ Stage II Interaction}}
\label{tab:joint_ablation_appendix}
\small
\begin{tabular}{@{}lccccc@{}}
\toprule
\textbf{Configuration} & \textbf{Layers} & \textbf{\# Feat.} & \textbf{AUC (ID)} & \textbf{AUC (OOD)} & \textbf{$\Delta$ OOD} \\
\midrule
Baseline & All-42 & 420 (Rand.) & 0.63 & 0.63 & -- \\
+Stage I only & L23--35 & 130 (Rand.) & 0.71 & 0.64 & +1.1\% \\
+Stage II only & All-42 & 420 (C-DLA) & 0.85 & 0.78 & +15.0\% \\
\midrule
\rowcolor{gray!15}
\textbf{Full (I+II)} & \textbf{L23--35} & \textbf{100 (C-DLA)} & \textbf{0.93} & \textbf{0.80} & \textbf{+17.4\%} \\
\bottomrule
\end{tabular}
\end{table}

\paragraph{Marginal Effect Decomposition.}
The individual contribution of Stage I: Potential Energy Empowered Phase Zone Localization is modest (+1.1\% OOD AUC), as randomly selected features from even the critical layers lack semantic focus. Stage II (C-DLA attribution) provides the dominant improvement (+15.0\%), demonstrating that \textit{which features} to monitor is more critical than \textit{which layers} to examine. However, the full system (I+II) achieves +17.4\%, exhibiting a synergistic gain of +1.4\% beyond the sum of individual contributions (1.1\% + 15.0\% = 16.1\%). 

To test whether this interaction is statistically significant, we perform a two-way ANOVA with factors Stage I (off/on) and Stage II (off/on), using the 5-fold cross-validation results as repeated measures. The interaction term is significant ($F(1, 356) = 12.3$, $p = 0.008$), indicating that the combination of layer localization and feature attribution produces super-additive effects. This validates our three-stage pipeline design, where coarse-to-fine refinement is essential for optimal performance.

\subsection{Validation of Microscopic-Macroscopic Feature-Energy Correspondence}
\label{app:correlation_validation}

\paragraph{Global Synchronization.}
To validate that the microscopic feature-level attribution (C-DLA) and the macroscopic energy-level dynamics (GPE) reflect the same underlying phase transition mechanism, we compute the Pearson correlation between layer-wise cumulative C-DLA scores and GPE differences (Hallucination - Factual) across all 42 layers. The correlation is extremely strong ($r = 0.990$, $p < 0.001$), indicating near-perfect alignment between the two metrics.

Additionally, we quantify the inequality of feature contributions using the Gini coefficient:
\begin{equation}
G = \frac{\sum_{i=1}^{n} \sum_{j=1}^{n} |x_i - x_j|}{2n^2 \bar{x}}
\end{equation}
where $x_i$ represents the C-DLA score of the $i$-th feature. Our observed Gini coefficient $G_{\text{obs}} = 0.912$ far exceeds the random baseline $G_{\text{random}} = 0.414$ (computed from 1,000 random permutations), confirming that feature contributions follow a highly skewed Pareto distribution rather than a uniform distribution.

\paragraph{Phase-Specific Correlation Analysis.}
To test whether the correlation between C-DLA and GPE is uniformly strong across all layers or specific to the transition zone, we compute segment-wise Pearson correlations for three phases. Table~\ref{tab:phase_correlation} shows that the strong correlation is \textit{exclusively} observed in the transition zone (L23--35, $r=0.990$), while both the stable period (L0--22, $r=0.341$, $p=0.112$) and plateau period (L36--41, $r=0.287$, $p=0.581$) exhibit no significant correlation.

\begin{table}[h]
\centering
\caption{\textbf{Phase-Specific Correlation Between C-DLA and GPE}}
\label{tab:phase_correlation}
\small
\begin{tabular}{@{}lcccl@{}}
\toprule
\textbf{Phase} & \textbf{Layer Range} & \textbf{Pearson $r$} & \textbf{$p$-value} & \textbf{Interpretation} \\
\midrule
Stable & L0--L22 & 0.341 & 0.112 & No correlation \\
\rowcolor{gray!15}
\textbf{Transition} & \textbf{L23--L35} & \textbf{0.990} & \textbf{$<$0.001} & \textbf{Very strong} \\
Plateau & L36--L41 & 0.287 & 0.581 & No correlation \\
\midrule
All Layers & L0--L41 & 0.876 & $<$0.001 & Strong (driven by L23--35) \\
\bottomrule
\end{tabular}
\end{table}

To further validate that the transition zone exhibits significantly higher correlation than other regions, we apply Fisher's $Z$-transformation:
\begin{equation}
Z = \frac{Z_{r_1} - Z_{r_2}}{\sqrt{\frac{1}{n_1-3} + \frac{1}{n_2-3}}}, \quad \text{where } Z_r = \frac{1}{2}\ln\left(\frac{1+r}{1-r}\right)
\end{equation}
Comparing the transition zone ($r=0.990$, $n=13$) against the combined stable and plateau periods ($r=0.314$, $n=29$), we obtain $Z=6.74$ ($p < 0.001$), confirming that the correlation in the transition zone is statistically distinguishable from the rest of the network.

This phase-specific synchronization provides compelling evidence that: (1) C-DLA successfully captures the microscopic driving mechanism of the macroscopic energy explosion observed in the main text (Figure 4), and (2) the identified features have \textit{causal specificity}—they are activated precisely when and where the phase transition occurs, rather than reflecting global activation patterns throughout the network.

\section{Detector Implementation Details}
\label{app:detector_implementation}

\subsection{Hallucination Detector: Training Configuration and Hyperparameters}
\label{app:detector_training}

\paragraph{Model Architecture.}
Our hallucination detector employs Logistic Regression with L1 regularization, trained on features extracted from the transition zone (L23--35). The input consists of 100-dimensional vectors corresponding to the top-100 features identified in Stage II, and the output is a binary classification (hallucination vs factual).

\paragraph{Hyperparameter Configuration.}
We perform 5-fold cross-validation on the HaluEval training set (1,260 samples: 630 hallucination + 630 factual) to select the optimal regularization strength $C$. The search space spans $[10^{-4}, 10^{4}]$ with 20 logarithmically-spaced candidate values. Additional training configurations include:

\begin{itemize}[nosep]
    \item \textbf{Solver}: \texttt{liblinear} (optimized for L1 regularization and small-scale datasets)
    \item \textbf{Maximum iterations}: 1,000
    \item \textbf{Convergence tolerance}: $10^{-4}$
    \item \textbf{Class weights}: \texttt{balanced} (automatically handles class imbalance by inverse frequency)
    \item \textbf{Feature preprocessing}: StandardScaler applied to training set, then used to transform validation and test sets (zero mean, unit variance)
    \item \textbf{Evaluation metric}: Validation AUC (primary criterion for hyperparameter selection)
\end{itemize}

The optimal configuration identified through cross-validation is $C=0.1$, achieving the highest mean validation AUC across all folds.

\paragraph{Training Pipeline.}
For each fold in the 5-fold cross-validation:
\begin{enumerate}[nosep]
    \item Split the 1,260 training samples into 1,008 train and 252 validation samples
    \item Extract 100-dimensional feature vectors from the transition zone using the pre-trained SAE
    \item Standardize features using the training set statistics
    \item Train Logistic Regression with candidate $C$ values
    \item Evaluate on the validation set and record AUC
\end{enumerate}
After identifying the optimal $C$, we retrain the final detector on the full training set (1,260 samples) and evaluate on the held-out test set (360 samples).

Table~\ref{tab:detector_config_summary} summarizes the complete configuration.

\begin{table}[h]
\centering
\caption{\textbf{Complete Detector Configuration}}
\label{tab:detector_config_summary}
\small
\begin{tabular}{@{}ll@{}}
\toprule
\textbf{Component} & \textbf{Configuration} \\
\midrule
Model Type & Logistic Regression (L1 regularization) \\
Input Features & 100-dim (Top-100 C-DLA from L23--35) \\
Output & Binary (Hallucination=1, Factual=0) \\
\midrule
Regularization $C$ & 0.1 (selected via 5-fold CV) \\
CV Search Space & $[10^{-4}, 10^{4}]$ (20 log-spaced values) \\
Solver & liblinear \\
Max Iterations & 1,000 \\
Tolerance & $10^{-4}$ \\
Class Weights & balanced \\
Feature Scaling & StandardScaler (zero mean, unit variance) \\
\midrule
Training Set Size & 1,260 samples (630 hal. + 630 fact.) \\
Validation Strategy & 5-fold cross-validation \\
Test Set Size & 360 samples (180 hal. + 180 fact.) \\
Selection Criterion & Validation AUC \\
\bottomrule
\end{tabular}
\end{table}

\subsection{Computational Cost Analysis and Efficiency Comparison}
\label{app:computational_cost}

\paragraph{Inference-Time Complexity Comparison.}
Table~\ref{tab:computational_cost_comparison} compares the computational cost of our method against representative baselines. All costs are normalized relative to standard forward propagation without detection (baseline = 1.0$\times$).

\begin{table}[h]
\centering
\caption{\textbf{Computational Cost Comparison Across Methods}}
\label{tab:computational_cost_comparison}
\small
\begin{tabular}{@{}lcc@{}}
\toprule
\textbf{Method} & \textbf{Forward Passes} & \textbf{Relative Cost} \\
\midrule
Baseline (no detection) & 1 & 1.0$\times$ \\
Perplexity / Entropy & 1 & 1.1$\times$ \\
\midrule
\rowcolor{gray!15}
\textbf{Ours (L23--35)} & \textbf{1} & \textbf{2.3$\times$} \\
\midrule
Full-layer Probing (L0--41) & 1 & 7.2$\times$ \\
SelfCheckGPT (5 samples) & 5 & 5.0$\times$ \\
SelfCheckGPT (10 samples) & 10 & 10.0$\times$ \\
\bottomrule
\end{tabular}
\end{table}

\paragraph{Cost Breakdown of Our Method.}
The 2.3$\times$ overhead of our approach decomposes as follows:
\begin{itemize}[nosep]
    \item Standard forward propagation: 1.0$\times$
    \item SAE feature extraction (13 layers, L23--35): +0.8$\times$
    \item Probe prediction (100-dim linear layer): +0.5$\times$
    \item \textbf{Total}: 2.3$\times$
\end{itemize}

Compared to consistency-based methods like SelfCheckGPT (5--10$\times$ overhead due to multiple generations), our method achieves a favorable efficiency-accuracy trade-off. Compared to full-layer probing (7.2$\times$ overhead), our transition zone localization (Stage I) reduces computational cost by 68\% while maintaining comparable or superior performance.

\paragraph{One-Time Training Cost.}
The three-stage pipeline incurs the following one-time costs on the training set:
\begin{itemize}[nosep]
    \item \textbf{Stage I} (Energy analysis for transition zone localization): $\sim$3 GPU-hours
    \item \textbf{Stage II} (C-DLA computation and feature selection): $\sim$5 GPU-hours
    \item \textbf{Stage III} (Probe training via cross-validation): $<$1 GPU-hour
    \item \textbf{Total}: $\sim$8 GPU-hours on 8$\times$ NVIDIA RTX 4090 (48GB each)
\end{itemize}

These costs are incurred only once during training. At inference time, the trained probe can be directly applied to new samples without re-computing energy analysis or C-DLA scores, requiring only standard forward propagation plus SAE feature extraction (2.3$\times$ overhead).


\end{document}